\documentclass[preprint,12pt]{elsarticle}
\usepackage{soul}
\usepackage{multirow}
\usepackage[table,xcdraw]{xcolor} %
\usepackage{float}
\usepackage{pgfplots}
\usepackage{url}
\usepackage{pdflscape}
\usepackage{setspace}
\usepackage{booktabs}
\usepackage{fancyvrb}

\usepackage{amssymb}

\journal{}

\begin{document}
\definecolor{ao}{rgb}{0.0, 0.5, 0.0}

\begin{frontmatter}



\title{Optimal Strategies to Perform Multilingual Analysis of Social Content for a Novel Dataset in the Tourism Domain}


\author[inst1,inst2]{Maxime~Masson\corref{c1}}
\cortext[c1]{Corresponding author}
\ead{maxime.masson@univ-pau.fr}

\affiliation[inst1]{organization={LIUPPA, E2S, University of Pau and Pays Adour (UPPA)}}

\author[inst2]{Rodrigo~Agerri}
\ead{rodrigo.agerri@ehu.eus}

\author[inst1]{Christian~Sallaberry}
\ead{christian.sallaberry@univ-pau.fr}

\author[inst1]{Marie-Noelle~Bessagnet}
\ead{marie-noelle.bessagnet@univ-pau.fr}

\author[inst1]{Annig~Le Parc~Lacayrelle}
\ead{annig.lacayrelle@univ-pau.fr}

\author[inst1]{Philippe~Roose}
\ead{philippe.roose@univ-pau.fr}

\affiliation[inst2]{organization={HiTZ Center - Ixa},
country={University of the Basque Country UPV/EHU}}

\begin{abstract}
The rising influence of social media platforms in various domains, including tourism, has highlighted the growing need for efficient and automated Natural Language Processing (NLP) strategies to take advantage of this valuable resource. However, the transformation of multilingual, unstructured, and informal texts into
structured knowledge still poses significant challenges, most notably the never-ending requirement 
for manually annotated data to train deep learning classifiers. In this work, we study different NLP techniques to establish the best ones to obtain competitive performances while keeping the need for
training annotated data to a minimum. To do so, we built the first publicly available multilingual dataset (French, English, and Spanish) for the tourism domain,
composed of tourism-related tweets. The dataset includes multilayered, manually revised annotations for
Named Entity Recognition (NER) for Locations and Fine-grained Thematic Concepts Extraction mapped to the Thesaurus of Tourism and Leisure Activities of the World Tourism Organization, as well as for Sentiment Analysis at the tweet level. Extensive experimentation comparing various few-shot and fine-tuning techniques with modern language models demonstrate that modern few-shot techniques allow us to obtain competitive results for all three tasks with very little annotation data: 5 tweets per
label (15 in total) for Sentiment Analysis, 30 tweets for Named Entity Recognition of Locations and 1K tweets annotated with fine-grained thematic concepts, a highly fine-grained sequence labeling task based on an inventory of 315 classes. We believe that our results, grounded in a novel dataset, pave the way for applying NLP to new domain-specific applications, reducing the need for manual annotations and circumventing the complexities of rule-based, ad-hoc solutions.
\end{abstract}




\end{frontmatter}



\section{Introduction}

Social media platforms have become essential channels for
sharing opinions and experiences about tourist practices and itineraries.
Twitter, in particular, has become a popular medium for people to spontaneously
share their thoughts and recommendations, making it a valuable source of
information and user-generated content (UGC) for the tourism industry
\cite{zeng2014we} (such as tourism offices, destination marketing organizations
- DMOs, etc.).  

However, analyzing vast volumes of social media data can be challenging
\cite{maynard2012challenges}, especially when it comes to extracting structured
knowledge from unstructured text. Consequently, tourism stakeholders often
delegate the task of knowledge extraction to researchers, who rely on Natural Language Processing (NLP) techniques. NLP offers a powerful set of techniques for processing and analyzing text data and is often used to address three
common knowledge extraction tasks: {Sentiment Analysis}, {NER for Locations}, and {Fine-grained Thematic Concept Extraction}
\cite{rosenthal-etal-2015-semeval,liu2022overview,fu2020clinical}.

Recently, NLP techniques based on deep learning and language models have emerged. These techniques offer several advantages over traditional rule-based ones. Deep learning-based techniques can adapt to changing language patterns and structures \cite{min2021recent}, ensuring a more
dynamic and up-to-date analysis. However, to achieve optimal results in domain-specific applications, language models must be fine-tuned. Consequently, researchers often face two recurrent challenges: (1) {determining which NLP technique is most suitable for a given domain} (e.g., which technique, which language model, etc.), and (2)
{discerning how many domain-specific examples are truly necessary to achieve competitive NLP results}. As annotating datasets is both costly and time-consuming, researchers strive to keep the annotation work to a minimum while maintaining high-quality results.

In this paper, we present a comparative study on the data requirements for achieving competitive performance in NLP tasks within the {tourism domain}. We hypothesize that among the existing language models and training approaches, some will be better suited for this specific domain, especially given the unique context of social media characterized by short informal texts, frequent grammatical errors, and the presence of emojis. We focus on the three common knowledge extraction tasks mentioned above: Sentiment Analysis (text
classification), NER for Locations (sequence labeling), and Fine-grained Thematic Concept Extraction (sequence labeling). Specifically, we investigate which NLP techniques are best to keep manual data annotation to a minimum and to avoid cumbersome and costly rule-based ad-hoc approaches. To enable this
study, we contribute a new manually annotated and multilingual (French, English, Spanish) dataset extracted from social media, which includes annotations for the three tasks mentioned above.

The main contributions are the following: (i) we present a {novel dataset} of tourism tweets. This dataset is multilingual (French, English, and Spanish) and has been manually annotated at the text level with sentiment labels and at the token level with locations and thematic concepts linked to a fine-grained tourism thesaurus (World Tourism Organization Thesaurus of Tourism and Leisure Activities \cite{World_Tourism_Organization2002-hs}) which makes it, to the best of our knowledge, the first of its kind. {The dataset is publicly available \footnote{https://huggingface.co/datasets/mx-phd/tourism}.}; (ii) we perform a {comparative study} on
rule-based, fine-tuning, and few-shot learning techniques with the aim of
establishing which of the techniques is the most efficient for each NLP task
(Sentiment Analysis, NER for Locations and Fine-grained Thematic Concept Extraction) on social media data for the tourism domain; (iii) additionally, we experiment with various numbers of examples and dataset sampling techniques to determine {how many annotated examples are really needed} to achieve competitive results for each task, using different training techniques and language models.

{Experiments with various Masked Language Models (encoder-only MLMs) and Large Language Models (LLMs, decoder and encoder-decoder) in few-shot and fine-tuning settings demonstrate that it is possible to obtain competitive results for all three tasks with very little annotation data: 5 tweets per label (15 in total) for Sentiment Analysis, 30 examples using generative LLMs for NER and 1000 tweets annotated with {thematic concepts}. Our results also suggest that few-shot prompting for sequence labeling \cite{ma-etal-2022-template} using MLMs seems to be particularly effective for highly fine-grained tasks (more than 300 classes). Overall, the obtained results indicate that MLMs applied in few-shot settings remain competitive with respect to LLMs for discriminative tasks. This is coherent with previous results published for this type of task \cite{setfit}.}

These results provide a promising solution to apply NLP to new domain-specific applications, keeping the manual annotation requirements low while avoiding complex rule-based ad-hoc solutions. The code and annotated data will be made publicly available to facilitate the reproducibility of the results and encourage research on this particular topic. In summary, we believe that the findings of our paper may be useful not only for researchers interested in the tourism domain but for any application that requires NLP analysis in a domain-specific scenario when no annotated data is available, while avoiding ad-hoc rule-based approaches.

The rest of the paper is structured as follows. In Section~\ref{sec:related-work},
we provide an overview of the NLP techniques based on deep learning used in tourism. Section~\ref{sec:dataset-building} covers the construction and annotation of the dataset. Section~\ref{sec:experimental-setup} describes the experimental setup.
In Section~\ref{sec:results}, we present a comparative analysis of fine-tuning and
few-shot techniques for three common NLP tasks. Furthermore, the
results and limitations of the experiment are discussed in Section~\ref{sec:discussion}. Finally,
Section~\ref{sec:conclusion} provides some concluding remarks and offers some insight into the future application of the work presented here. 

\section{Related Work} \label{sec:related-work}

{In the rapidly evolving field of Natural Language Processing (NLP), one of the most significant advancements has been the advent of pre-trained language models based on the Transformer architecture {\cite{vaswani2017attention}}. These models are trained on vast corpora, capturing a broad spectrum of linguistic structures, nuances, and knowledge \cite{min2021recent,Manning2020EmergentLS,toporkov2023role}. As a result, they offer a significant boost in performance and generalization for every NLP task. In this paper, we use all three types of language models based on the Transformer architecture.}


\begin{itemize}
    \item {\emph{Masked Language Models} (we refer to them as MLMs), use the {encoder} block of the Transformer \cite{vaswani2017attention}. The learning objective of MLMs consists of learning to predict masked words from the surrounding context. Popular models include BERT (Bidirectional Encoder Representations from Transformers) \cite{DBLP:journals/corr/abs-1810-04805} or XLM-RoBERTa \cite{xlm-roberta}.}
    \item {\emph{Large Language Models} (LLMs) they are text-to-text models based on both blocks (encoder-decoder) or only the decoder component of the Transformer. These models are generative and, while their most successful results have come in text generation tasks, they have also started to be used for discriminative tasks in few-shot settings \cite{chung2024scaling,garcia2024medical,sainz2024gollie}. Generative LLMs include the {GPT} ({Generative Pre-Trained Transformer}) series of models \cite{brown2020language}, Mistral \cite{jiang2023mistral}, LLaMa 2 ({Large Language Model Meta AI}) series \cite{touvron2023llama2} or Google's FlanT5 \cite{chung2024scaling} to name but a few.}
\end{itemize}

\subsection{Previous Work on Social Media for the Tourism Domain}

A popular approach in NLP is to {fine-tune} language models for domain-specific downstream tasks (refer to {Table~\ref{tab:overviewFineTune}}). This
results in altering the model weights to adapt it to the new task. For instance, language models have been fine-tuned for text classification tasks, including spam detection in hotel reviews \cite{crawford2021using} and Sentiment Analysis in touristic reviews \cite{enriquez2022transformers,vasquez2021bert} or reviews about sustainable transport \cite{su13042397}, leading to improved accuracy. In Named Entity Recognition (NER), fine-tuning of language models has been employed to extract
location information from a tourism corpus \cite{bouabdallaoui2022named,cheng2020mtner}. Furthermore, language models have demonstrated
promising results in thematic concept extraction, such as identifying travel-related themes and topics from tourism texts
\cite{chantrapornchai2021information}.



\begin{table}[H]
\centering
  \begin{tabular}{r|c|l|l}
  \toprule
   \textbf{Ref.} & \textbf{Year} & \textbf{Objective} & \textbf{Label} \\
   \midrule
    \cite{crawford2021using} & 2021 & Spam Detection (Hotel Reviews) & Text \\ 
    \cite{enriquez2022transformers} & 2022 & Sentiment Analysis (Tourism Reviews)  & Text \\ 
    \cite{vasquez2021bert} & 2021 & Sentiment Analysis  (Tourism Reviews) & Text \\ 
    \cite{su13042397} & 2021 & Sentiment Analysis (Transport) & Text  \\ 
    \cite{fernandez2018gazteak} & 2018 & Classification of Basque Users & Text  \\ 
    \cite{bouabdallaoui2022named} & 2022 & Location Extraction (Touristic Corpus) & Token \\ 
    \cite{cheng2020mtner} & 2020 & Location Extraction (Touristic Corpus) & Token \\ 
    \cite{chantrapornchai2021information} & 2021 & Travel Themes Identification & Token \\
    \bottomrule
    
    \end{tabular}
    \caption{Examples of Application of Fine-Tuning of Language Models in the Tourism Domain.}
    \label{tab:overviewFineTune}

\end{table}

The main limitation of fine-tuning language models is that, in most cases, it requires {substantial amount of manually annotated data to obtain competitive results} \cite{DBLP:journals/corr/abs-1810-04805}. These datasets are not always available, and sometimes they require being built from scratch, a costly and time-consuming process that researchers strive to avoid.

\subsection{Addressing the Lack of Domain-Specific Annotated Data}
\label{sec:lackdata}

\emph{Zero-shot} and \emph{few-shot} learning techniques have emerged as effective approaches to mitigate the need for manually annotated training data. Thus, instead of fine-tuning the pre-trained model's weights to a downstream task, prompting the language models in zero and few-shot settings allows us to obtain competitive results in classification tasks \cite{kadam2020review}. 

\subsubsection{{Few-Shot with Generative Large Language Models (LLMs)}}

{In the case of generative LLMs (such as GPT {\cite{brown2020language}}, Mistral {\cite{jiang2023mistral}} or LLaMa 2 {\cite{touvron2023llama2}}), it is possible to apply them in zero-shot by simply describing the task to be carried out in natural language. Most commonly, they are generation tasks, but they can also be prompted to perform text classification and sequence labeling tasks such as Sentiment Analysis and NER, respectively. Furthermore, sometimes adding a few examples may help, as illustrated by the following 2-shot prompt for Sentiment Analysis.}

\begin{Verbatim}[commandchars=\\\{\}]
{You are an assistant that classifies sentiments of texts.}
{You must classify them as: positive, negative, or neutral.}
{Examples:}
{User: "We went to the beach yesterday, it was amazing!"}
{Assistant: positive}
{User: "So bad, it's raining today. Have to stay home ... :("}
{Assistant: negative}
{User: "Beautiful sun today"}
{Assistant: ...}
\end{Verbatim}

{Few-shot prompting is interesting, especially in scenarios in which domain-specific annotated data is rare and has been applied with promising results {\cite{brown2020language}}. However, results across domains are mixed. For example, studies have found that it can perform poorly in some domains, like the biomedical one {\cite{moradi2021gpt}}.}

\subsubsection{{Few-Shot with Masked Language Models (MLMs)}}

{With respect to {text classification} tasks, Pattern-Exploiting Training (PET) is a semi-supervised few-shot training approach that uses MLMs as a backbone. It combines the idea of providing the MLM with task descriptions in natural language and a cloze-style phrase generation approach to help the model understand the task {\cite{schick2020exploiting}}. For example, to classify movie reviews based on the predominant sentiment they express, the model would be prompted with the query: \emph{The movie was $\langle MASK \rangle$}. The model would then try to predict the $\langle MASK \rangle$, choosing from options such as outstanding ({positive}) or terrible ({negative}).} 

{More recently, SetFit {\cite{setfit}} provides a prompt-free framework for few-shot fine-tuning of {Sentence Transformers}. It leverages {contrastive learning}, where only a small number of labeled examples are needed to fine-tune a pre-trained model. {\cite{setfit}}. SetFit attains high accuracy using minimal labeled data. For example, it requires just eight labeled examples per class on the customer reviews sentiment dataset to be competitive with fine-tuned RoBERTa-large {\cite{DBLP:journals/corr/abs-1907-11692}} on the full training set of 3k examples {\cite{setfit}}.}

{Regarding {sequence labeling} tasks, several recent studies have also explored new approaches to replace complex templates used in few-shot prompting, such as for NER \cite{wang2022instructionner,ma-etal-2022-template}. For example, EntLM ({Entity-oriented LM}) \cite{ma-etal-2022-template} aims to simplify the process of generating task-specific queries and reduce the reliance on manual template construction. EntLM currently represents the state-of-the-art for few-shot NER.}

\subsection{Cross-lingual Transfer Techniques}


{An alternative to few-shot learning to address the lack of annotated data for NLP tasks on social media is the use of
multilingual language models to perform {data augmentation via machine translation} or {crosslingual model-transfer}. In the first case, the idea is to translate into multiple languages the annotated training data from a source
language and then use the translated versions to perform data augmentation during training. This technique has been tested for many sequence labeling \cite{garcia-ferrero-etal-2022-model,garciaferrero2022tprojection,yeginbergen2024crosslingual} and classification tasks \cite{artetxe2020translation,artetxe2023revisiting} in various genres of text, including Sentiment Analysis in
social media \cite{barriere-balahur-2020-improving}. In the second case, the idea is to leverage the multilingual capabilities of some MLMs and LLMs to learn in a source language and predict in a different target language. Although interesting,
cross-lingual transfer techniques are based on transferring {existing}
annotations from (at least) a source to a given target language(s). However, our
starting point is the absence of {any annotated data in any language} for touristic locations, {fine-grained touristic concepts} and sentiment for the tourism domain.}

\subsection{Existing Annotated Resources}

While publicly available annotated data for the tourism domain is non-existent, there are several annotated corpora from other domains that could be used for
experimentation, as highlighted in Table~\ref{tab:dataset-details}. Here, we compare a selection of
existing annotated datasets along different criteria: (1) the source of the
data collection, (2) the types of annotations available, (3) the languages
covered by the dataset, and (4) the methodology used for generating annotations
(manual by humans, semi-automatic or automatic).

For example, the {ESTER} corpus \cite{ester2006} is a comprehensive
collection of French radio transcripts and {AnCora} \cite{ancora2008},
is a multilevel annotated corpus ({mostly from newspaper}) for Catalan
and Spanish. Both of these resources are annotated for named entities
({such as persons, locations, organizations}). 

In terms of social
media-specific resources, the {Broad Twitter Corpus} (BTC)
\cite{derczynski-etal-2016-broad} includes coarse-grained NER annotations,
while {Sentiment140} \cite{go2009twitter}, {STS-Gold}
\cite{saif2013evaluation}, and many other datasets developed as part of shared
evaluation tasks at {SemEval}
\cite{nakov-etal-2013-semeval,rosenthal-etal-2014-semeval,rosenthal-etal-2015-semeval,
nakov-etal-2016-semeval, rosenthal-etal-2017-semeval}, are used for sentiment
analysis. Other corpora include the {MultiWOZ} dialogue dataset
\cite{budzianowski-etal-2018-multiwoz}, the {Stanford NLI} dataset
\cite{bowman2015large} for text inference, and the {Heldugazte} corpus
\cite{fernandez2018gazteak}, which assists in categorizing tweets as
formal or informal.

\begin{table}[H]
    \centering

    \begin{tabular}{r|l|l|c|l}
            \toprule
        \textbf{Dataset} & \textbf{Source} & \textbf{Annotations}  & \textbf{Lang.} & \textbf{Type} \\
        \midrule
        ESTER \cite{ester2006} & Radio & Named Entities & FR & Manual \\
        AnCora \cite{ancora2008} & Newspapers & Named Entities & ES, CA & Semi \\
        BTC \cite{derczynski-etal-2016-broad}  & Twitter & Named Entities & EN & Manual \\
        Sent140 \cite{go2009twitter} & Twitter & Sentiment & EN & Auto \\
        STS-Gold \cite{saif2013evaluation} & Twitter & Sentiment & EN & Manual \\
        GSC \cite{su13042397} & Reviews & Sentiment & EN & Auto \\
        SemEval \cite{nakov-etal-2013-semeval} & Twitter & Sentiment & EN & Manual \\
        MultiWOZ \cite{budzianowski-etal-2018-multiwoz} & Humans & Dialogue States & EN & Manual \\
        SNLI \cite{bowman2015large} & Humans & Inference Pairs & EN & Manual \\
        Heldugazte \cite{fernandez2018gazteak} & Twitter & Formal/Informal & EU & Auto \\
        \bottomrule
    \end{tabular}
    \caption{Comparison of Existing Annotated Datasets for Various NLP Tasks}
    \label{tab:dataset-details}
\end{table}

These datasets are extensive but broad and often focus on English only,
therefore lacking the necessary contextual information relevant to the tourism
domain. Most importantly, we could not find any public dataset annotated for
{Fine-grained Thematic Concept Extraction} in the tourism domain. Taking this into account, we decided to build our own custom annotated dataset.

\section{Dataset Building and Annotation}\label{sec:dataset-building}

In this section, we describe the creation of a novel multilingual dataset\footnote{{https://huggingface.co/datasets/mx-phd/tourism}} consisting of tourism-related tweets annotated for three important NLP tasks for tourist applications: (1) {Sentiment Analysis}, (2) {Named Entity
	Recognition for Locations}, and (3) {Fine-grained Thematic Concept
Extraction} (based on the World Tourism Organization Thesaurus on Tourism and Leisure Activities
\cite{World_Tourism_Organization2002-hs}). 

\subsection{Collection}

The dataset was collected from Twitter using the Academic
API\footnote{https://developer.twitter.com/en/use-cases/do-research/academic-research (discontinued in April 2023)}
and the collection process was carried out by applying a novel methodology that we
have been designed for dataset building. This methodology is both generic,
iterative and incremental \cite{masson2022domain}. 
Several iterations were carried out, each of them with successive
filtering corresponding to the {dimensions} of the target dataset: 

\begin{itemize}
    \item \emph{Spatial}: {French Basque Coast} area ({defined by spatial coordinates for geotagged tweets or a list of toponyms}). 
    \item \emph{Temporal}: Summer of 2019 ({21 Jun -- 21 Sept}). Based on tweets' timestamps. 
    \item \emph{Thematic}: Tourism domain ({defined by the World Tourism Organization Thesaurus on Tourism and Leisure}). 
\end{itemize}

To prevent excessive noise, each iteration was followed by human feedback to adjust and balance the filters. For example, a large number of tweets related
to the G7 summit, which took place in the region that year, were collected
because they contained tourism-related keywords; these were subsequently
blacklisted using hashtag and keyword exclusion rules. Similarly, we excluded
professional and institutional users because we are primarily interested in
understanding the behaviors and feelings of individual tourists, not analyzing
promotional or institutional content.

{The data collection strictly adhered to Twitter’s Academic Research API policy, which, at the time of collection, authorized the retrieval of up to 10 million tweets per month for non-commercial research. In accordance with GDPR and standard ethical practices, only tweet texts and user IDs were retained; all personal metadata, including usernames and profile information, were removed to ensure user anonymization and privacy protection.}

The final dataset is made up of 27,379 tweets, of which 2,961 tweets from 624
users were selected to be annotated and used for experimentation. The tweets in
this subset (2,961) were manually checked to confirm that they were
{about tourism} and {from tourist visitors} (unlike the tweets from
tourism professionals, or from news outlets speaking about tourism, etc.).  The
dataset is multilingual and includes an unbalanced variety of tweets in
English, French, and Spanish, which reflects the reality of the use of social
media on the {French Basque Coast}. Table~\ref{datasetStats}
provides the language distribution of the dataset and the splits created for
experimentation (60\% train, 20\% dev, 20\% test). Splits were generated
maintaining a balance between the number of users and languages in each set.

\begin{table}[H]
\centering

\begin{tabular}{r|c|c|c|c}
\toprule
       & \textbf{All} & \textbf{French}  & \textbf{English}  & \textbf{Spanish} \\ 
       \midrule
\textbf{Train}  & 1,662 (503) & 1,297 (391) & 283 (129) & 82 (32)  \\ 
\textbf{Dev}  & 619 (300) & 450 (213) & 99 (66)  & 70 (31)  \\ 
\textbf{Test}  & 680 (431) & 401 (273) & 102 (100) & 177 (93) \\ 
\bottomrule
\end{tabular}
\caption{Breakdown of the Collected Dataset by Language -- Tweets (Users)}
\label{datasetStats}
\end{table}



\subsection{Sentiment Annotation}

\begin{figure}[H]
    \centering
    \includegraphics[width=.8\linewidth]{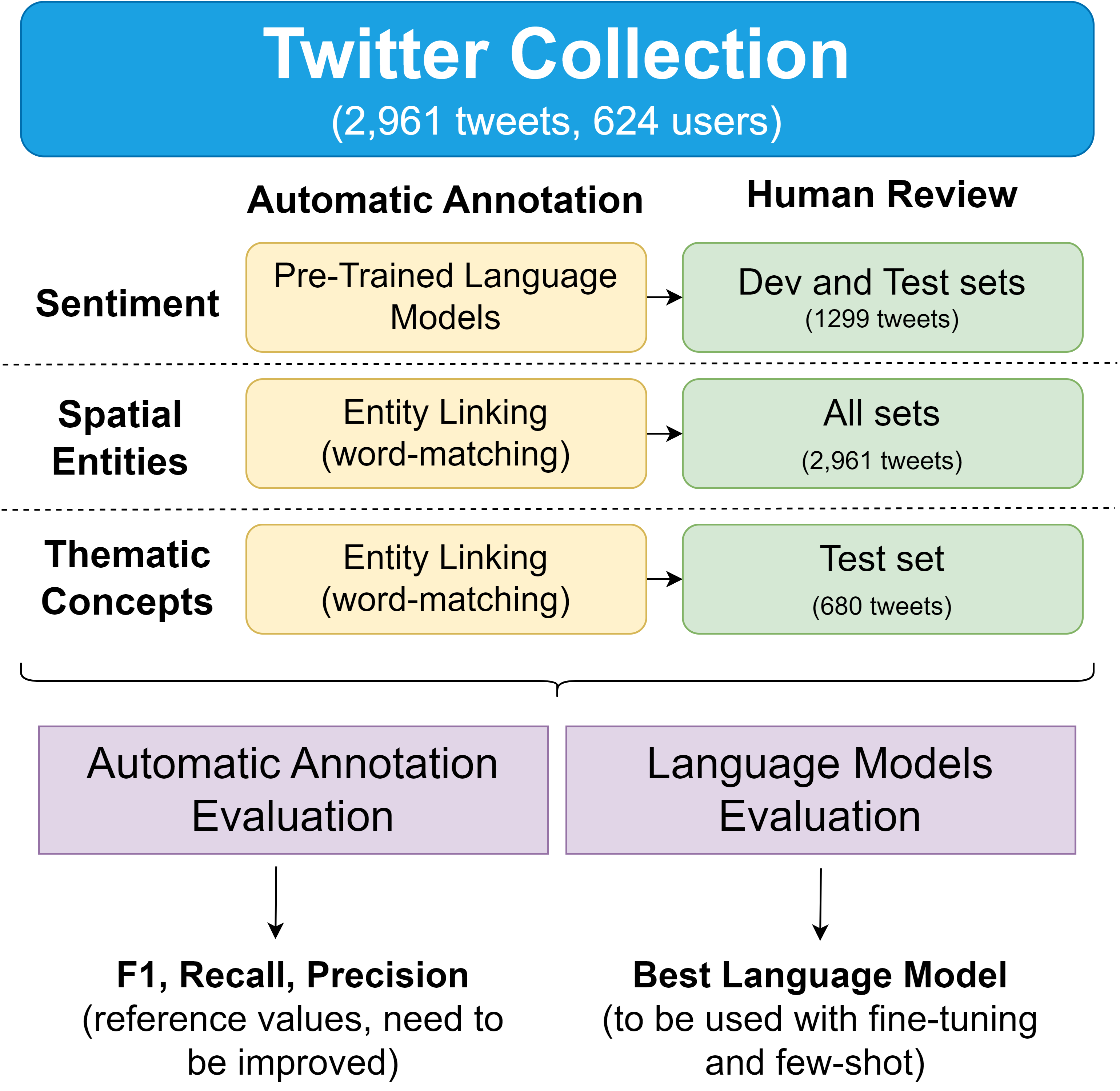}
    \caption{Dataset Building and Annotation Process}
 \label{fig:datasetBuildingProcess}
\end{figure}

The annotation process of those 2,961 tweets was carried out in a
semi-automatic manner, following the procedure depicted in
Figure~\ref{fig:datasetBuildingProcess}. 

Firstly, to assist human annotators, the 1,299 tweets in the
development and test splits underwent a process of
{automatic annotation} using the 5 language models listed in Table
\ref{tab:accuracy-sentiment}. Subsequently, they were manually reviewed. 
Each tweet was assigned to two annotators to evaluate agreement
({Cohen's kappa coefficient}) and ensure the quality of the annotations. We
achieved $\kappa = 0.79$ for French tweets, $\kappa = 0.75$ for Spanish tweets,
and $\kappa = 0.67$ for English tweets, which corresponds to a strong agreement.
Any disagreements were resolved through collaborative discussion.

The next step was to evaluate the performance of the five language models used to
automatically label the tweets with respect to the human annotations. Table~\ref{tab:accuracy-sentiment} shows that
XLM-T Sentiment, fine-tuned with multilingual Sentiment Analysis data from various domains different from tourism \cite{barbieri2022xlm}, outperformed, on average for the three languages, any
other approaches. Following this, we automatically annotated using XLM-T Sentiment the training split, which was then manually revised for its use in the
experimentation.

\begin{table}[H]
\centering

  \begin{tabular}{r|ccccc}
  \toprule
   \textbf{Sentiment Models:}  
   & \rotatebox{90}{{Barbieri et al. (2020) \cite{barbieri2020tweeteval}}} 
   & \rotatebox{90}{Pérez and Carlos (2021) {\cite{perez2021pysentimiento}}}  
   & \rotatebox{90}{Seethal (2022) {\cite{seethal}}}
   & \rotatebox{90}{Hartmann et al. (2023) {\cite{hartmann2023}}} 
   & \rotatebox{90}{Barbieri et al. (2022) {\cite{barbieri2022xlm}}} \\
   \midrule
    
    ~ & \multicolumn{5}{c}{\textbf{French}} \\ 
        \textbf{Global} & 0.56 & 0.45 & 0.43 & 0.47 & \textbf{0.82} \\
    \textbf{\textcolor{teal}{Positive}} & 0.34 & 0.14 & 0.11 & 0.95 &  0.82 \\
    \textbf{\textcolor{red}{Negative}} & 0.06 & 0.11 & 0.00 & 0.28 & 1.00 \\
    \textbf{\textcolor{blue}{Neutral}} & 0.97 & 0.97 & 1.00 & 0.00 & 0.74 \\
    
    \hline
   ~ & \multicolumn{5}{c}{\textbf{Spanish}} \\ 
    \textbf{Global} & 0.71 & 0.64 & 0.61 & 0.34 & \textbf{0.83} \\
    \textbf{\textcolor{teal}{Positive}} & 0.31 & 0.09 & 0.03 & 1.00 & 0.84 \\
    \textbf{\textcolor{red}{Negative}} & 0.00 & 0.00 & 0.00 & 0.29 & 0.43 \\
    \textbf{\textcolor{blue}{Neutral}} & 1.00 & 1.00 & 0.98 & 0.00  & 0.87 \\
    
    \hline
    ~ & \multicolumn{5}{c}{\textbf{English}} \\ 
    \textbf{Global} & \textbf{0.81} & \textbf{0.81} & 0.71 & 0.66 & 0.80 \\
    \textbf{\textcolor{teal}{Positive}} & 0.75 & 0.75 & 0.59 & 1.00 & 0.72 \\
    \textbf{\textcolor{red}{Negative}} & 0.75 & 0.50 & 0.75 & 0.50 & 0.75 \\
    \textbf{\textcolor{blue}{Neutral}} & 0.94 & 0.97 & 0.94 & 0.00 & 0.97 \\
    \bottomrule
  \end{tabular}
    \caption{Accuracy of Available Sentiment Language Models on Manually
  Annotated Test Data}
  \label{tab:accuracy-sentiment}
\end{table}

\subsection{Locations and Thematic Concepts}

Although Sentiment Analysis is a text classification task in which each tweet is
assigned a polarity label, we are also interested in identifying locations and
thematic concepts relevant to the tourism domain. These two tasks are
addressed as sequence labeling problems.

Before experimenting with supervised techniques, we implemented a basic rule-based \emph{word-matching} approach as a baseline for {locations} and
{thematic concepts}. Locations were matched using 625 local toponyms
extracted from Open Street Map ({cities, POIs, landmarks, etc.}) while thematic concepts were matched using their label and synonyms in the WTO thesaurus of tourism (which contains 1,494 touristic concepts). 

Tweet preprocessing (lowercase, removal of URLs, hashtag splitting, decomposing hashtags to
find concepts or toponyms in them) was performed to facilitate
{word-matching}. We applied this algorithm to annotate the train, dev, and test
splits. Automatic annotations were then manually corrected by human annotators: for
{locations}, all train, dev, and test sets. For {thematic
concepts} the \emph{word-matching} algorithm detected 315 concept classes for the full
dataset (out of the 1,494 concepts included in the WTO ontology), making it a highly
fine-grained sequence labeling task. Due to this fact, we just revised the
{thematic concepts} for the test set, as annotating 315 concept classes is a
complex task requiring a large human effort. Finally, inter-annotator agreement was calculated on a subset of 100 random tweets. For
location entities: $\kappa = 0.91$ for exact matches, when all tokens forming
an entity are precisely the same (e.g., New (B-LOC), - (I-LOC), York (I-LOC)). On the other hand, $\kappa = 0.93$ for
partial matches, when an entity is mostly recognized but has missing or extra
tokens (e.g., New (B-LOC), - (O), York (O)). Both values indicate a near-perfect consensus.

The F1-score results of evaluating \emph{word-matching} baseline on the test set are reported
in Table~\ref{tab:perfEL}. These results will serve as a baseline to compare with
the different supervised techniques in the experimental section. 

\begin{table}[H]
\centering
\begin{tabular}{cccc}
\toprule
\multicolumn{4}{c}{\textbf{Named Entity Recognition (NER) for Locations}} \\
\midrule
& \multicolumn{1}{c}{\textbf{Recall}} & \multicolumn{1}{c}{\textbf{Precision}} & \multicolumn{1}{c}{\textbf{F1-score}}  \\ 
\multicolumn{1}{r|}{Location Exact Match} & \multicolumn{1}{c|}{0.692} & \multicolumn{1}{c|}{0.722} & 0,707 \\ 
\multicolumn{1}{r|}{Location Partial Match} & \multicolumn{1}{c|}{0.780} & \multicolumn{1}{c|}{0.814} & 0,796 \\ 
\midrule
\multicolumn{4}{c}{\textbf{Fine-grained Thematic Concept Extraction}} \\
\midrule
\multicolumn{1}{c}{} & \multicolumn{1}{c}{\textbf{Recall}} & \multicolumn{1}{c}{\textbf{Precision}} & \multicolumn{1}{c}{\textbf{F1-score}} \\ 
\multicolumn{1}{r|}{Concept Exact Match} & \multicolumn{1}{c|}{0.746} & \multicolumn{1}{c|}{0.952} & 0,836 \\  
\multicolumn{1}{r|}{Concept Partial Match} & \multicolumn{1}{c|}{0.747} & \multicolumn{1}{c|}{0.953} & 0,837 \\
\bottomrule
\end{tabular}
\caption{Performance of the \emph{word-matching} Algorithm on both Sequence Labeling Tasks}
\label{tab:perfEL}
\end{table}

From the results in Table \ref{tab:perfEL}, it can be observed that, for
locations, results are not that good, particularly in terms of recall. However,
the \emph{word-matching} algorithm performs remarkably well on Fine-grained
Thematic Concept Extraction, especially in terms of precision. 
Although this constitutes a rather strong baseline, the recall remains comparatively low,
which means that many thematic concepts remain undetected by the system.
Furthermore, developing such as \emph{word-matching} algorithm is a rather complex and time-consuming exercise which we would ideally like to avoid for any domain-specific new application.

Thus, our main objective is now to establish whether deep-learning supervised
techniques based on multilingual language models can match or improve over the
{word-matching} algorithm while keeping the amount of manual annotation to
a minimum, especially for {Fine-grained Thematic Concept Extraction}. Thus, in addition
to standard fine-tuning techniques, it is of particular interest to investigate
techniques based on few-shot learning, where the aim is to generate competitive
taggers using only a very small amount of labeled data. 

\begin{table}[H]
  \centering

  \begin{tabular}{r|c|c|c|c}
  \toprule
   \textbf{Set} & \textbf{Tweets} & \textbf{Locations} & \textbf{Concepts} & \textbf{Sentiments} \\ 
   \midrule
    {Train} & 1,662 & 4,030 & 3,841 & \textcolor{teal}{787} (+) \textcolor{red}{191} (-) {684} (=)\\ 
    {Dev} & 619 & 1,419 & 1,337 & \textcolor{teal}{271} (+) \textcolor{red}{82} (-) {266} (=)\\ 
    {Test} & 680 & 1,679 & 1,844 & \textcolor{teal}{299} (+) \textcolor{red}{93} (-) {288} (=)\\ 
    \textbf{Global} & 2,961 & 7,128 & 7,022 & \textcolor{teal}{1,357} (+) \textcolor{red}{366} (-) {1,238} (=)\\
    \bottomrule
  \end{tabular}
  \caption{Dataset Annotations ({following human review})}
  \label{tab:annotationsDataset}
\end{table}

The data used for experimentation is the one generated by manually revising the
annotations for the three tasks, as described above.
Table~\ref{tab:annotationsDataset} provides a detailed description of the
dataset, including the total number of tweets (2,961) and the number of annotations
per task. Note that for the sentiment column in Table~\ref{tab:annotationsDataset}, the colors and symbols used represent each type of polarity (\textcolor{red}{{-}} for negative annotations, \textcolor{teal}{{+}} for positive and {{=}} for neutral).


In summary, the goal is to establish how much labeled data we actually
need to obtain competitive performance by comparing the {fine-tuning} and
{few-shot} techniques with respect to the \emph{word-matching} algorithm,
and which of these techniques is the most efficient in terms of performance and
human effort.

\section{Experimental Setup}\label{sec:experimental-setup}

{Figure~\ref{fig:comparative}} provides an overview of our experimental setup with the language models, sampling techniques, and learning techniques used for each task. The experimentation is focused on the three tasks presented above: {Sentiment Analysis}
(see Section~\ref{sec:setupText}), {Named Entity Recognition (NER) for Locations} and
{Fine-grained Thematic Concept Extraction} (see Section~\ref{setupToken}). {These experiments leverage the tourist dataset described in the previous section ({2,961
multilingual tweets including 1,662 for training}) and summarized in Table
{\ref{tab:annotationsDataset}}. 

\begin{figure}[H]
    \centering
    \includegraphics[width=1\linewidth]{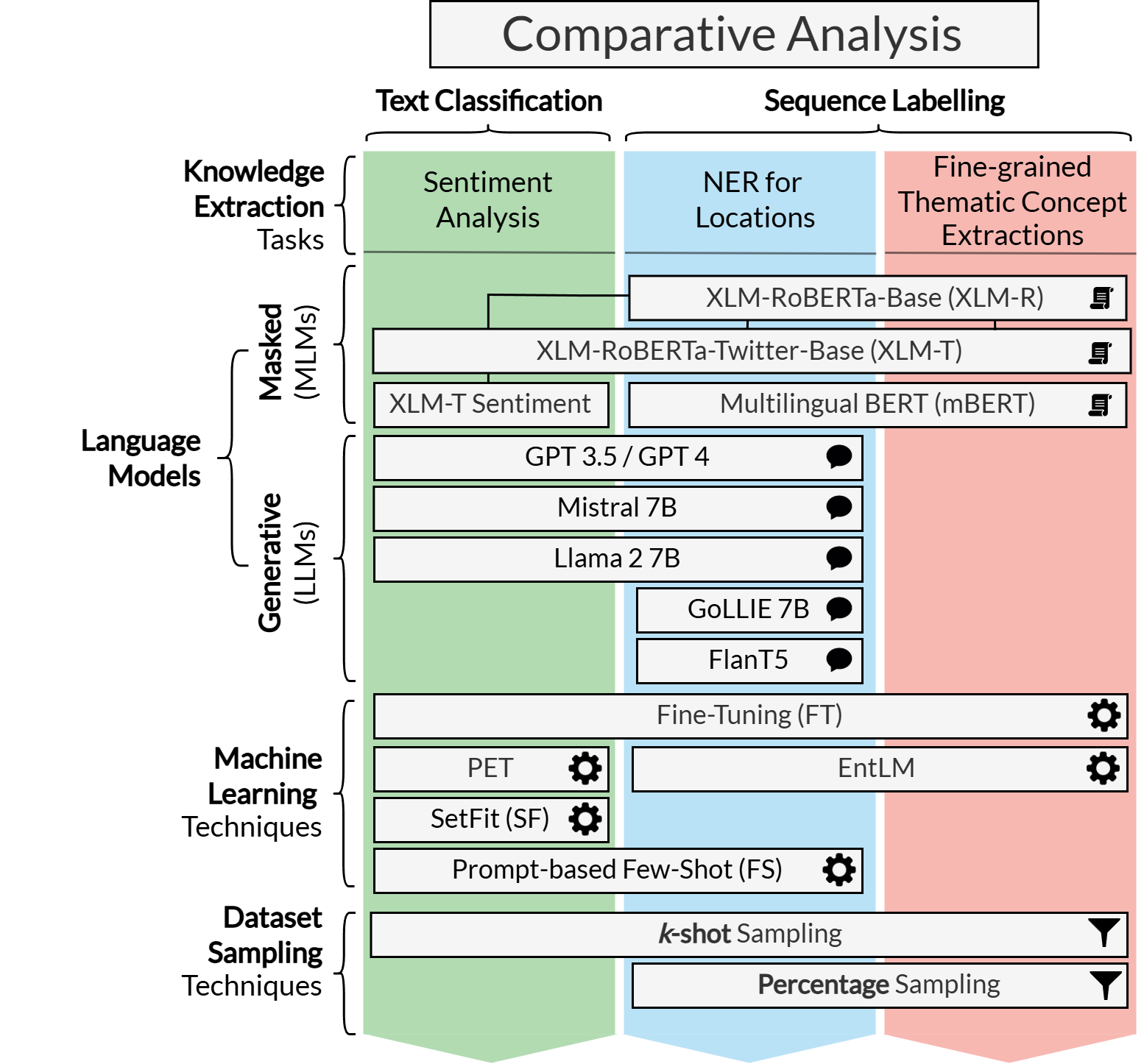}
    \caption{{Experimental Setup of the Comparative Study}}
    \label{fig:comparative}
\end{figure}

To study machine learning approaches, the dataset is sampled using two different techniques (see {Figure~\ref{fig:comparative}}, \emph{Dataset Sampling Techniques}):}

\begin{itemize}
\item \emph{$k$-shot sampling}: In this technique, we selected a specific number
of examples for each tweet or token label from the training set. We
	performed training or prompting using the following k-values: 5, 10, 20, 30, 40, 50, and
100 examples per label. For Sentiment Analysis, we used the PET $k$-shot sampling
technique \cite{schick2020exploiting} while for {locations} and
	{fine-grained thematic concepts}, we apply the EntLM $k$-shot technique
with default parameters \cite{ma-etal-2022-template}.
\item \emph{Percentage  sampling}: For sequence labeling ({locations}
	and {fine-grained thematic concepts}) we also experiment with sampling
on percentages of tweets rather than labels, as $k$-shot does. We
	successively used 5\%, 10\%, 20\%, 30\%, 40\%, 50\%, 60\%, 70\%, 90\%, and
	100\% of the training set, while trying to maintain the original label
	classes distribution, including the O labels (the O label is assigned to tokens which are neither location nor thematic entities), which the $k$-shot sampling technique does not contemplate.
\end{itemize}


\subsection{Text Classification -- Sentiment Analysis}\label{sec:setupText}

Firstly, based on the results reported in Table~\ref{tab:accuracy-sentiment}, we used MLMs (see {Figure~\ref{fig:comparative}}, \emph{Language Models, Masked}). We chose the XLM-T for experiments on Sentiment Analysis. XLM-T is based on XLM-RoBERTa \cite{xlm-roberta}, further pre-trained on a corpus of 198 million tweets for 15 languages \cite{barbieri2022xlm}. This version of the model is specifically designed to handle the unique characteristics of tweets and social media posts, such as their limited length, informal language, and the presence of emojis and hashtags. More specifically, we use two variants of the XLM-T model:

\begin{itemize}
    \item The {base version} \cite{barbieri2022xlm} (XLM-T).
	\item XLM-T was previously fine-tuned specifically for
		Sentiment Analysis \cite{barbieri2022xlm} (XLM-T Sentiment). This sentiment variant has already been fine-tuned using 24,264 out-of-domain tweets in eight different languages, including French, English, and Spanish. However, it is important to note that these tweets cover a wide range of topics, which do not necessarily include tourism.
\end{itemize}

{We also experiment with the following generative LLMs (see Figure{~\ref{fig:comparative}}, \emph{Language Models, Generative}):}

\begin{itemize}
    \item {GPT 3.5 {\cite{brown2020language}} (gpt-3.5-turbo-0125)\footnote{\url{https://platform.openai.com/docs/models/gpt-3-5-turbo}}, which is the latest version of GPT 3.5 with improved instruction following. This model is paid and closed source, and was used through the OpenAI API\footnote{\url{https://openai.com/blog/openai-api}}. Additionally, we also use GPT 4 (gpt-4-0125-preview) but only in zero-shot settings due to API cost limitations. For both models, we use the default {temperature} of 1 and a {presence penalty} of 0.} 
    
    \item {Mistral 7B {\cite{jiang2023mistral}}: We use Mistral-7B-Instruct-v0.2, the 7 billion parameters instruct version of the model. A {batch size} of 1, {learning rate} of 2e-4, and {weight decay} of 0.001 are used, as recommended in \cite{kaggleFinetuneMistral}. } 
    
    \item {LLaMa 2 7B {\cite{touvron2023llama2}}: Similar to Mistral, we experiment with the 7 billion parameters instruct version of the model, namely, Llama-2-7b-chat-hf. The same hyperparameters as for Mistral are used, which were also recommended in \cite{kaggleFinetuneLlama}.} 
\end{itemize}

{LLMs (e.g., Mistral 7B, LLaMa 2 7B, GPT 3.5, and GPT 4), are applied in zero-shot and few-shot settings (FS), prompting the model with a few examples. We use the Sentiment Analysis prompt presented in Section~{\ref{sec:lackdata}}. In the case of Mistral 7B and LLaMa 2 7B, as those models are open-source, we also experiment with fine-tuning them for Sentiment Analysis. We use the techniques and hyperparameter settings introduced in previous similar works {\cite{kaggleFinetuneMistral,kaggleFinetuneLlama}}.}

With respect to the two MLMs (e.g., XLM-T and XLM-T Sentiment), we used them as a backbone for three different training techniques (see Figure{~\ref{fig:comparative}}, \emph{Machine Learning Techniques}) along various dataset sizes using the sampling techniques described previously.

\begin{itemize}
    \item \emph{Fine-Tuning} (FT): optimal hyperparameters were found through grid search. For XLM-T 8 batch size, 2e-5 learning rate, weight
		decay 0.01; for XLM-T Sentiment 32 batch, 1e-5 learning rate
		and decay: 0.1.
	\item \emph{Pattern-Exploiting Training} (PET)
		\cite{schick2020exploiting} is used with default hyperparameters.
\item {\emph{SetFit} (SF) {\cite{setfit}}: a prompt-free framework for few-shot fine-tuning of {sentence transformers}. We use the training parameters recommended in the SetFit repository\footnote{\url{https://github.com/huggingface/setfit}}, namely 4 epochs of training with a batch size of 16.}
\end{itemize}

By comparing these deep learning techniques and assessing their effectiveness with varying amounts of annotated data, we aim to learn any insights about the minimum data requirements for achieving reliable Sentiment Analysis results in the tourism domain. This would allow us to establish which technique requires less annotated data to obtain competitive performance.

\subsection{Sequence Labeling -- Locations and Fine-grained Thematic Concept Extraction}
\label{setupToken}

{For the sequence labeling task of NER for Locations, we experiment with and evaluate the following techniques.} 

\begin{itemize}
    \item {\emph{Zero- and Few-Shot Sequence Labeling with Generative LLMs} (FS): we use the same generative LLMs as for Sentiment Analysis to have a common reference point, namely GPT 3.5, GPT 4, Mistral 7B, and LLaMa 2 7B}.
    
    \item {\emph{EntLM} {\cite{ma-etal-2022-template}} with a multilingual BERT (mBERT){\cite{DBLP:journals/corr/abs-1810-04805}} as backbone. The hyperparameters used are those recommended in the EntLM repository, namely a {batch size} of 4, {learning rate} of 1e-4, and {weight decay} of 0. } 
    
    \item {For \emph{Fine-Tuning} (FT) with MLMs, in addition to XLM-T, we include two additional models: XLM-RoBERTa {\cite{xlm-roberta}} (XLM-R) and the previously mentioned mBERT. Grid search for hyperparameter tuning found as optimal values 8 for batch size, 5e-5 for learning rate, and 0.1 for weight decay. To fine-tune generative LLMs for sequence labeling, we leverage the library published by García-Ferrero et al. {\cite{garcia2024medical}}. This fine-tuning technique allows performing sequence labeling tasks as a text-to-text generation task. We experiment with two generative models: LLaMa 2 7B and FlanT5 (more specifically, flan-t5-base).
    }
    
    \item {\emph{GoLLIE} ({Guideline following Large Language Model for Information Extraction}): a specialized language model trained to follow annotation guidelines {\cite{sainz2024gollie}}. It allows the user to perform sequence labeling inferences based on annotation schemes. The GoLLIE architecture can be enriched using domain-specific training examples. We will leverage the GoLLIE model with its base configuration of 7 billion parameters, pairing it with our training dataset. }
\end{itemize}

{In our analysis of Fine-grained Thematic Concept Extraction, we focused on two primary approaches: (1) employing the EntLM framework and (2) the Fine-Tuning (FT) of MLMs.} {Alternative approaches such as {few-shot learning with generative LLMs} and GoLLIE proved impractical due to the extensive number of thematic classes (e.g., 315 classes) involved. The sheer volume of classes exceeded the context window capacity of current models, leading to errors or the generation of random text unrelated to the task.}

\begin{table}[H]
  \centering
  {
  \begin{tabular}{l|c|c|c}
  \toprule
   \textbf{Technique} & \textbf{Batch} & \textbf{Learning Rate} & \textbf{Decay} \\ 
   \midrule
    \multicolumn{4}{c}{\textbf{Text Classification}}\\
    PET & \multicolumn{3}{c}{Default (Schick et al., 2020) \cite{schick2020exploiting}} \\ 
    SetFit (SF) & 16 & \multicolumn{2}{c}{4 epochs} \\ 
    Fine-Tuning (XLM-T Sent) & 32 & 1e-5 & 0.1 \\ 
    Fine-Tuning (Mistral/LLaMa 2) & 1 & 2e-4 & 0.001 \\ 
    
    \multicolumn{4}{c}{\textbf{Sequence Labeling}}\\
    EntLM (mBERT) & 4 & 1e-4 & 0 \\ 
    GoLLIE & \multicolumn{3}{c}{Default (Sainz et al., 2024) \cite{sainz2024gollie}}  \\ 
    Fine-Tuning (XLM-R/mBERT) & 8 & 5e-5 & 0.1 \\ 
    Fine-Tuning (FlanT5/LLaMa 2) & \multicolumn{3}{c}{As per García-Ferrero et al. (2024) \cite{garcia2024medical}} \\

    \multicolumn{4}{c}{\textbf{Both}}\\
    Fine-Tuning (XML-T) & 8 & 2e-5 & 0.01 \\ 
    Few-Shot (GPT-3.5/GPT-4) & \multicolumn{3}{c}{Temp: 1, Presence Penalty: 0} \\ 
    Few-Shot (Mistral/LLaMa 2) & \multicolumn{3}{c}{Default \cite{touvron2023llama2} \cite{jiang2023mistral}} \\ 
    
    \bottomrule
  \end{tabular}}
  \caption{{Overview of the Hyperparameters used for each Approach}}
  \label{tab:hyperparameters}
\end{table}

Moreover, we are interested in comparing the results of these approaches with the baseline established by the \emph{word-matching} (rules-based) algorithm, as presented in Table \ref{tab:perfEL}. The primary objective of this comparison is to ascertain the minimum amount of annotated data required to justify transitioning from costly and manual \emph{word-matching} approaches
to more advanced deep learning techniques for each task, respectively. More specifically, we seek to determine the tipping point at which the benefits of employing deep learning techniques outweigh their data requirements. This is particularly true for a highly complex task such as {Fine-grained Thematic Concept Extraction}, which involves labeling 315 different concept classes, and for which developing \emph{word-matching} algorithms or manually annotating data are highly inefficient and expensive approaches.

{To handle noise in tweets (e.g., spelling errors, slang), we rely on the inherent robustness of pretrained language models, which have been trained on large-scale, noisy corpora including social media text. These models generalize well to informal language, spelling variations, and slang, eliminating the need for additional noise-specific preprocessing. Their learned representations implicitly normalize such variability. For related work demonstrating language models' robustness to noise, see \cite{brown2020language} and \cite{hendrycks2020measuring}.}

Experiments were conducted on servers equipped with Nvidia A100 (80GB VRAM) GPUs, Intel Xeon Gold 6226R CPUs (2.90 GHz) and 256 GB of RAM, and language models were accessed from the HuggingFace Transformers API \cite{wolf-etal-2020-transformers}.

As it is customary, for Sentiment Analysis, we report accuracy results, while for sequence labeling we use the usual F1-micro metric calculated at the span level as defined in the CoNLL 2002 shared task \cite{tjong-kim-sang-2002-introduction}. All reported results are the average of three randomly initialized runs.

\section{Results}\label{sec:results}

We apply the experimental setup presented in the previous section to our multilingual dataset of tweets from the tourism domain.

\subsection{Sentiment Analysis}
\definecolor{darkred}{rgb}{0.55, 0.0, 0.0}

{The results of the Sentiment Analysis on the five techniques are reported in Table~{\ref{tab:resultSentimentTable}}. As a reminder, this task consists of classifying the polarity of each tweet as {positive}, {negative}, or {neutral}. We have highlighted in \textbf{bold} the results we will refer to in the text.}

\begin{table}[H]
\centering
\scriptsize

{\setstretch{1.75}
\begin{tabular}{l|ccccccccc|}
\cline{2-10}
 &
  \multicolumn{9}{c|}{\textbf{Examples per class} (positive, negative and neutral) — \textbf{Accuracy}} \\ \hline
\multicolumn{1}{|c|}{\textbf{Techniques}} &
  \multicolumn{1}{c|}{\textbf{0}} &
  \multicolumn{1}{c|}{\textbf{5}} &
  \multicolumn{1}{c|}{\textbf{10}} &
  \multicolumn{1}{c|}{\textbf{20}} &
  \multicolumn{1}{c|}{\textbf{30}} &
  \multicolumn{1}{c|}{\textbf{40}} &
  \multicolumn{1}{c|}{\textbf{50}} &
  \multicolumn{1}{c|}{\textbf{100}} &
  \multicolumn{1}{c|}{\textbf{All}} \\ \hline
\multicolumn{1}{|l|}{\textbf{{Prompt-based FS}}} &
  \multicolumn{9}{c|}{\cellcolor[HTML]{E7E6E6}{Regular \textbf{Prompt-based Few-Shot} of LLMs}} \\ \hline
\multicolumn{1}{|l|}{{GPT 3.5}} &
  \multicolumn{1}{c|}{\cellcolor[HTML]{8FCC85}\textbf{{0.785}}} &
  \cellcolor[HTML]{9CD087}0.739 &
  \cellcolor[HTML]{97CF86}0.757 &
  \cellcolor[HTML]{94CE86}0.766 &
  \cellcolor[HTML]{A9D48A}0.694 &
  \cellcolor[HTML]{ABD58B}0.685 &
  \cellcolor[HTML]{B1D78C}\textcolor{black}{{0.664}} &
  \cellcolor[HTML]{B6D98D}0.645 &
  \cellcolor[HTML]{E7E6E6} \\ \cline{1-2}
\multicolumn{1}{|l|}{{Mistral 7B}} &
  \multicolumn{1}{c|}{\cellcolor[HTML]{A2D289}\textbf{{0.716}}} &
  \cellcolor[HTML]{94CE86}0.766 &
  \cellcolor[HTML]{95CE86}0.764 &
  \cellcolor[HTML]{98CF87}0.754 &
  \cellcolor[HTML]{96CE86}0.761 &
  \cellcolor[HTML]{96CE86}0.760 &
  \cellcolor[HTML]{96CE86}\textcolor{black}{{0.758}} &
  \cellcolor[HTML]{96CE86}0.760 &
  \cellcolor[HTML]{E7E6E6} \\ \cline{1-2}
\multicolumn{1}{|l|}{{LLaMa 2 7B}} &
  \multicolumn{1}{c|}{\cellcolor[HTML]{F0EA99}0.442} &
  \cellcolor[HTML]{C6DD90}0.589 &
  \cellcolor[HTML]{C4DD90}0.598 &
  \cellcolor[HTML]{ACD58B}0.680 &
  \multicolumn{4}{c}{\cellcolor[HTML]{E7E6E6}{Exceeding Input Context   Length}} &
  \multirow{-4}{*}{\cellcolor[HTML]{E7E6E6}\emph{\begin{tabular}[c]{@{}c@{}}\end{tabular}}} \\ \hline
\multicolumn{1}{|l|}{\textbf{FT of MLMs}} &
  \multicolumn{9}{c|}{\cellcolor[HTML]{E7E6E6}Fine-Tune of \textbf{Encoder-Only Models} (MLMs)} \\ \hline
\multicolumn{1}{|l|}{XLM-T} &
  \multicolumn{1}{c|}{\cellcolor[HTML]{E7E6E6}} &
  \multicolumn{1}{c|}{\cellcolor[HTML]{F3EC9A}0.428} &
  \multicolumn{1}{c|}{\cellcolor[HTML]{FFEF9C}0.385} &
  \multicolumn{1}{c|}{\cellcolor[HTML]{DEE595}0.503} &
  \multicolumn{1}{c|}{\cellcolor[HTML]{D2E193}0.545} &
  \multicolumn{1}{c|}{\cellcolor[HTML]{BDDB8E}0.622} &
  \multicolumn{1}{c|}{\cellcolor[HTML]{B6D88D}\textbf{{0.646}}} &
  \multicolumn{1}{c|}{\cellcolor[HTML]{8DCC84}0.792} &
  \multicolumn{1}{c|}{\cellcolor[HTML]{77C580}\textcolor{black}{{0.868}}} \\ \cline{1-1} \cline{3-10} 
\multicolumn{1}{|l|}{XLM-T Sent} &
  \multicolumn{1}{c|}{\multirow{-2}{*}{\cellcolor[HTML]{E7E6E6}}} &
  \multicolumn{1}{c|}{\cellcolor[HTML]{6AC07D}\textbf{{0.917}}} &
  \multicolumn{1}{c|}{\cellcolor[HTML]{63BE7B}\textbf{{0.939}}} &
  \multicolumn{1}{c|}{\cellcolor[HTML]{68C07D}0.922} &
  \multicolumn{1}{c|}{\cellcolor[HTML]{75C47F}0.877} &
  \multicolumn{1}{c|}{\cellcolor[HTML]{76C47F}0.875} &
  \multicolumn{1}{c|}{\cellcolor[HTML]{67C07C}0.925} &
  \multicolumn{1}{c|}{\cellcolor[HTML]{6BC17D}\textcolor{black}{{0.914}}} &
  \multicolumn{1}{c|}{\cellcolor[HTML]{69C07D}{\textbf{0.919}}} \\ \hline
\multicolumn{1}{|l|}{{\textbf{FT of LLMs}}} &
  \multicolumn{9}{c|}{\cellcolor[HTML]{E7E6E6}{Fine-Tune of \textbf{Encoder-Decoder and Decoder-Only Models} (LLMs)}} \\ \hline
\multicolumn{1}{|l|}{{Mistral 7B}   } &
  \multicolumn{1}{c|}{\cellcolor[HTML]{E7E6E6}} &
  \multicolumn{1}{c|}{\cellcolor[HTML]{B8D98D}0.640} &
  \multicolumn{1}{c|}{\cellcolor[HTML]{BEDB8F}0.618} &
  \multicolumn{1}{c|}{\cellcolor[HTML]{BBDA8E}0.628} &
  \multicolumn{1}{c|}{\cellcolor[HTML]{A5D389}0.706} &
  \multicolumn{1}{c|}{\cellcolor[HTML]{99CF87}0.750} &
  \multicolumn{1}{c|}{\cellcolor[HTML]{A5D389}0.706} &
  \multicolumn{1}{c|}{\cellcolor[HTML]{93CD86}0.771} &
  \multicolumn{1}{c|}{\cellcolor[HTML]{83C882}0.828} \\ \cline{1-1} \cline{3-10} 
\multicolumn{1}{|l|}{{LLaMa 2 7B}   } &
  \multicolumn{1}{c|}{\multirow{-2}{*}{\cellcolor[HTML]{E7E6E6}}} &
  \multicolumn{1}{c|}{\cellcolor[HTML]{C5DD90}0.594} &
  \multicolumn{1}{c|}{\cellcolor[HTML]{B5D88D}0.651} &
  \multicolumn{1}{c|}{\cellcolor[HTML]{BFDB8F}0.613} &
  \multicolumn{1}{c|}{\cellcolor[HTML]{9CD087}0.738} &
  \multicolumn{1}{c|}{\cellcolor[HTML]{95CE86}0.763} &
  \multicolumn{1}{c|}{\cellcolor[HTML]{96CE86}0.759} &
  \multicolumn{1}{c|}{\cellcolor[HTML]{98CF87}\textbf{{0.761}}} &
  \multicolumn{1}{c|}{\cellcolor[HTML]{7EC781}\textcolor{black}{{0.844}}}  \\ \hline
\multicolumn{1}{|l|}{\textbf{PET}} &
  \multicolumn{9}{c|}{\cellcolor[HTML]{E7E6E6}{\textbf{Cloze-Style Few-Shot} with MLMs}} \\ \hline
\multicolumn{1}{|l|}{XLM-T} &
  \multicolumn{1}{c|}{\cellcolor[HTML]{E7E6E6}} &
  \multicolumn{1}{c|}{\cellcolor[HTML]{D6E294}0.533} &
  \multicolumn{1}{c|}{\cellcolor[HTML]{C1DC8F}0.607} &
  \multicolumn{1}{c|}{\cellcolor[HTML]{B2D78C}0.661} &
  \multicolumn{1}{c|}{\cellcolor[HTML]{A9D48A}0.691} &
  \multicolumn{1}{c|}{\cellcolor[HTML]{A1D288}0.722} &
  \multicolumn{1}{c|}{\cellcolor[HTML]{95CE86}\textbf{{0.764}}} &
  \multicolumn{1}{c|}{\cellcolor[HTML]{8CCB84}0.796} &
  \multicolumn{1}{c|}{\cellcolor[HTML]{74C47F}0.880} \\ \cline{1-1} \cline{3-10} 
\multicolumn{1}{|l|}{XLM-T   Sentiment} &
  \multicolumn{1}{c|}{\multirow{-2}{*}{\cellcolor[HTML]{E7E6E6}}} &
  \multicolumn{1}{c|}{\cellcolor[HTML]{C4DD90}0.598} &
  \multicolumn{1}{c|}{\cellcolor[HTML]{A2D289}0.717} &
  \multicolumn{1}{c|}{\cellcolor[HTML]{9FD188}0.729} &
  \multicolumn{1}{c|}{\cellcolor[HTML]{85C983}0.819} &
  \multicolumn{1}{c|}{\cellcolor[HTML]{8ECC85}0.787} &
  \multicolumn{1}{c|}{\cellcolor[HTML]{7BC681}0.855} &
  \multicolumn{1}{c|}{\cellcolor[HTML]{76C47F}0.874} &
  \multicolumn{1}{c|}{\cellcolor[HTML]{75C47F}0.877} \\ \hline
\multicolumn{1}{|l|}{\textbf{{SetFit (SF)}}} &
  \multicolumn{9}{c|}{\cellcolor[HTML]{E7E6E6}{Combination of \textbf{Few-Shot and Fine-Tuning} for Sentence Transformers}} \\ \hline
  \cline{1-1} \cline{3-10} 
\multicolumn{1}{|l|}{{XLM-T}} &
  \multicolumn{1}{c|}{\cellcolor[HTML]{E7E6E6}} &
  \multicolumn{1}{c|}{\cellcolor[HTML]{D6E294}0.534} &
  \multicolumn{1}{c|}{\cellcolor[HTML]{C8DE91}0.582} &
  \multicolumn{1}{c|}{\cellcolor[HTML]{A4D389}\textcolor{black}{{0.712}}} &
  \multicolumn{1}{c|}{\cellcolor[HTML]{A3D289}0.715} &
  \multicolumn{1}{c|}{\cellcolor[HTML]{91CD85}0.776} &
  \multicolumn{1}{c|}{\cellcolor[HTML]{9ED188}0.732} &
  \multicolumn{1}{c|}{\cellcolor[HTML]{8ACB84}0.803} &
  \multicolumn{1}{c|}{\cellcolor[HTML]{82C882}\textcolor{black}{{0.832}}} \\ \cline{1-1} \cline{3-10} 
\multicolumn{1}{|l|}{{XLM-T Sent.}} &
  \multicolumn{1}{c|}{\multirow{-3}{*}{\cellcolor[HTML]{E7E6E6}}} &
  \multicolumn{1}{c|}{\cellcolor[HTML]{82C882}0.831} &
  \multicolumn{1}{c|}{\cellcolor[HTML]{75C47F}0.878} &
  \multicolumn{1}{c|}{\cellcolor[HTML]{75C47F}\textcolor{black}{{0.876}}} &
  \multicolumn{1}{c|}{\cellcolor[HTML]{71C37E}0.893} &
  \multicolumn{1}{c|}{\cellcolor[HTML]{73C47F}0.882} &
  \multicolumn{1}{c|}{\cellcolor[HTML]{6FC27E}0.899} &
  \multicolumn{1}{c|}{\cellcolor[HTML]{7AC680}0.858} &
  \multicolumn{1}{c|}{\cellcolor[HTML]{85C983}0.821} \\ \hline
\end{tabular}}
\caption{{Sentiment Analysis with $k$-shot Sampling - Results on Text Classification Techniques (results in {\textbf{bold}} are referenced in the text)}}
\label{tab:resultSentimentTable}
\end{table}

{The most noteworthy aspect from the results is that fine-tuning XLM-T Sentiment (Table~{\ref{tab:resultSentimentTable}}, \emph{Fine-Tune of MLMs}) outperforms any other method using only 5 examples for training (Table~{\ref{tab:resultSentimentTable}}, {{0.919}}). In contrast to previous work {\cite{schick2020exploiting}}, this suggests that fine-tuning on a large multilingual dataset for Sentiment Analysis, even with texts from different domains, dramatically helps improve the results in domain-specific tourist data, clearly outperforming few-shot prompting techniques with MLMs such as PET {\cite{schick2020exploiting}} or LLMs. In fact, the fine-tuned XLM-T Sentiment model reaches optimal results with as few as 10 examples (Table~{\ref{tab:resultSentimentTable}}, {{0.939}})}.

{Among the techniques that use only our domain-specific training data, Mistral 7B obtains the best scores with only 5 examples (PET with XLM-T requires 50 examples to obtain a similar score, while SetFit performs similarly with 40-shot).}

{Concerning the methods using MLMs with only some examples from the training data, SetFit consistently outperforms fine-tuning until we reach 100 examples (e.g., Table~{\ref{tab:resultSentimentTable}}), but with more data, results from PET and fine-tuning are eventually better. Still, this highlights the effectiveness of SetFit, which can achieve high accuracy with only 40 examples and without requiring as many computing resources as for few-shot learning with LLMs.}

{
The performance of large language models (LLMs) in the zero-shot setting (Table~{\ref{tab:resultSentimentTable}}, \emph{Prompt-based Few-Shot}) shows that GPT 3.5 and Mistral 7B achieve competitive accuracy, with results ranging from 0.716 for Mistral 7B to 0.785 for GPT 3.5. In our experiments, GPT 3.5 outperformed GPT 4, achieving an accuracy of 0.785 compared to 0.757 for GPT 4. Due to the high cost associated with the GPT 4 API, higher shot configurations were not tested, and as a result, GPT 4’s performance in those settings is not reported in Table~\ref{tab:resultSentimentTable}.}

{
 The observed difference between GPT 3.5 and 4 may be attributed to the different underlying pre-training and post-training optimizations. Moreover, the performance of Mistral 7B, though slightly lower than GPT 3.5, is still competitive, which suggests that open-source models can achieve near-state-of-the-art results while being more cost-efficient.
}

\begin{figure}[H]
\centering
\begin{tikzpicture}
\begin{axis}[
    xlabel={Number of Examples},
    ylabel={Accuracy},
    xmin=0, xmax=100,
    ymin=0.3, ymax=0.9,
    legend cell align={left},
    ytick={0.1,0.2,0.3,0.4,0.5,0.6,0.7,0.8,0.9,1.0},
    xtick={10,20,30,40,50,60,70,100},
    xticklabels={5,10,20,30,40,50,100,All},       
    ymajorgrids=true,
    xmajorgrids=true,
    grid style=dashed,
    width=1\textwidth,
    legend pos=south east,
    legend columns=1,
]

\addlegendimage{empty legend}


\addplot[
    color=ao,
    mark=diamond*,
]
    coordinates {
    (0,0.716)
    (10,0.766)
    (20,0.764)
    (30,0.754)
    (40,0.761)
    (50,0.760)
    (60,0.758)
    (70,0.760)
};

\addplot[
    color=blue,
    mark=diamond*,
]
    coordinates {
    (10,0.533)
    (20,0.607)
    (30,0.661)
    (40,0.691)
    (50,0.722)
    (60,0.764)
    (70,0.796)
    (100,0.880)
};


\addplot[
    color=red,
    mark=diamond*,
]
    coordinates {
    (10,0.428)
    (20,0.385)
    (30,0.503)
    (40,0.545)
    (50,0.622)
    (60,0.646)
    (70,0.792)
    (100,0.868)
};


\addplot[
    color=orange,
    mark=diamond*,
]
    coordinates {
    (10,0.640)
    (20,0.618)
    (30,0.628)
    (40,0.706)
    (50,0.750)
    (60,0.706)
    (70,0.771)
    (100,0.828)
};


\addplot[
    color=purple,
    mark=diamond*,
]
    coordinates {
    (10,0.534)
    (20,0.582)
    (30,0.712)
    (40,0.715)
    (50,0.776)
    (60,0.732)
    (70,0.803)
    (100,0.832)
};


\draw [black, dashed, line width=1pt] (100,580) circle [radius=0.3cm];
\draw [black, dashed, line width=1pt] (100,520) circle [radius=0.3cm];
\node[font=\bfseries] at (93,580) {(b)};
\node[font=\bfseries] at (93, 520) {(c)};

\draw [black, dashed, line width=1pt] (30, 460) ellipse (4.5cm and 0.4cm);

\node[font=\bfseries] at (25,520) {(a)};


\addlegendentry{\hspace{0.8cm}\textbf{Techniques}}
\addlegendentry{{FS (Mistral 7B)}}
\addlegendentry{PET (XLM-T)}
\addlegendentry{FT of MLMs (XLM-T)}
\addlegendentry{{FT of LLMs (Mistral 7B)}}
\addlegendentry{{SetFit (XLM-T)}}


\end{axis}
\end{tikzpicture}
\caption{{Sentiment Analysis - $k$-shot Sampling}}
\label{fig:sentimentAnalysis}
\end{figure}
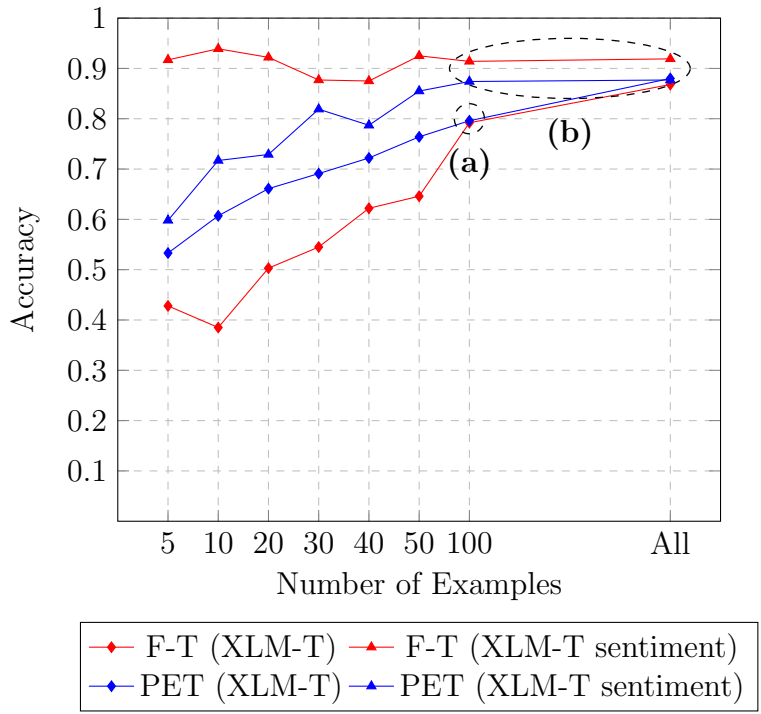

{Figure~{\ref{fig:sentimentAnalysis}} provides another perspective on the results. Here, the {x-axis} represents the number of training examples used (in this case, tweets), while the {y-axis} indicates the accuracy scores. As we have observed previously, XLM-T Sentiment outperforms other models in most configurations because it has prior knowledge of the task. In Figure{~\ref{fig:sentimentAnalysis}}, we focus on the other models (namely, XLM-T and  Mistral 7B) to analyze what would be the best technique in a text classification task where there is no available model with prior training like XLM-T Sentiment.}

{The observed patterns in Figure~{\ref{fig:sentimentAnalysis}} and Table~{\ref{tab:resultSentimentTable}} suggest the following four conclusions about Sentiment Analysis in the Tourism Domain:}
\begin{enumerate}
    \item {When using an already fine-tuned sentiment model such as XLM-T Sentiment (see Table~{\ref{tab:resultSentimentTable}}, \emph{Fine-Tune with MLMs} and XLM-T Sentiment), a dataset containing as few as 10 examples is sufficient to achieve state-of-the-art Sentiment Analysis performance in the tourism domain.}
    \item {When employing a base MLM such as XLM-T, SetFit appears to be a preferable choice in the tourism domain for few-shot scenarios, given that close to optimal performance can be reached with 40 examples per class (refer to Figure~{\ref{fig:sentimentAnalysis}}, 40 examples).}
    \item {In use cases where very little annotated data is available (e.g., less than 30 examples per class) and no language model fine-tuned for the task exists (like XLM-T Sentiment), few-shot with LLMs such as Mistral 7B is the best approach. This approach is able to produce an accuracy of roughly 0.750 consistently from 5 to 100 examples (refer to Figure~{\ref{fig:sentimentAnalysis}}, (a)). It also produces robust results in zero-shot settings (accuracy of 0.716).}
    \item {When no language model fine-tuned for the task exists, but a sizable amount of annotated data is available, Pattern-Exploiting Training and Fine-Tuning of MLMs (refer to Figure~{\ref{fig:sentimentAnalysis}}, (b)) produce slightly better results in the tourism domain than SetFit and Fine-Tuning of LLMs (refer to Figure~{\ref{fig:sentimentAnalysis}}). Additionally, in contexts where annotated data are widely available, MLMs still produce the best results.}
\end{enumerate}

{We will discuss the broader impact of these results later in Section{~\ref{sec:discussion}}. In the next section, we present the results obtained for the sequence labeling tasks, namely NER for Locations and Fine-grained Thematic Concept Extraction.}

\subsection{Named Entity Recognition (NER) for Locations}

{Table~{\ref{tab:resultNerTab}} reports the results of NER for Locations. When the full training dataset is employed, all techniques yield comparable F1-scores (refer to Table~{\ref{tab:resultNerTab}}, \emph{All Examples}).}

{GoLLIE obtains the best overall result with a 0.832 in F1-score using the full training data. In zero-shot GoLLIE, Mistral 7B and GPT 3.5 perform quite similarly, between 0.670 and 0.694 in F1-score. While GPT 3.5's few-shot scores are slightly higher, Mistral is an open-weights model. Thus, annotating only 20 to 30 examples should be enough to obtain competitive results with Mistral 7B.}

{Fine-tuning of MLMs with the full dataset produced F1-scores ranging from {{0.791}} with XLM-R to {{0.818}} with mBERT, and {{0.808}} with XLM-T (see Table~{\ref{tab:resultNerTab}}). Notably, mBERT slightly outperforms the XLM model series in this task. However, fine-tuning requires a substantial volume of labeled data, as evidenced by low F1-scores when it has seen only a few examples.}

\begin{table}[H]
\centering
\scriptsize
{
\setstretch{1.75}
\begin{tabular}{l|ccccccccc}
\cline{2-10}
 &
  \multicolumn{9}{c|}{\textbf{Examples per class} (location) — \textbf{F1-score}} \\ \hline
\multicolumn{1}{|c|}{\textbf{Techniques}} &
  \multicolumn{1}{c|}{\textbf{0}} &
  \multicolumn{1}{c|}{\textbf{5}} &
  \multicolumn{1}{c|}{\textbf{10}} &
  \multicolumn{1}{c|}{\textbf{20}} &
  \multicolumn{1}{c|}{\textbf{30}} &
  \multicolumn{1}{c|}{\textbf{40}} &
  \multicolumn{1}{c|}{\textbf{50}} &
  \multicolumn{1}{c|}{\textbf{100}} &
  \multicolumn{1}{c|}{\textbf{All}} \\ \hline
\multicolumn{1}{|l|}{\textbf{{Prompt-based FS}}} &
  \multicolumn{9}{c|}{\cellcolor[HTML]{E7E6E6}{Regular \textbf{Prompt-based Few-Shot} of LLMs}} \\ \hline
\multicolumn{1}{|l|}{{GPT 3.5}} &
  \multicolumn{1}{c|}{\cellcolor[HTML]{7DC781}\textbf{{0.694}}} &
  \cellcolor[HTML]{7DC681}0.698 &
  \cellcolor[HTML]{71C37E}0.762 &
  \cellcolor[HTML]{71C37E}0.762 &
  \cellcolor[HTML]{6AC17D}0.798 &
  \cellcolor[HTML]{68C07C}0.809 &
  \cellcolor[HTML]{64BF7C}0.828 &
  \multicolumn{1}{c|}{\cellcolor[HTML]{68C07D}0.806} &
  \multicolumn{1}{c|}{\cellcolor[HTML]{E7E6E6}} \\ \cline{1-9}
 \multicolumn{1}{|l|}{{Mistral 7B}   } &
  \multicolumn{1}{c|}{\cellcolor[HTML]{80C782}\textbf{{0.680}}} &
  \multicolumn{1}{c|}{\cellcolor[HTML]{7BC681}0.704} &
  \multicolumn{1}{c|}{\cellcolor[HTML]{7EC781}0.689} &
  \multicolumn{1}{c|}{\cellcolor[HTML]{77C580}0.730} &
  \multicolumn{1}{c|}{\cellcolor[HTML]{73C37F}\textbf{{0.749}}} &
  \multicolumn{1}{c|}{\cellcolor[HTML]{75C47F}0.741} &
  \multicolumn{1}{c|}{\cellcolor[HTML]{74C47F}0.742} &
  \multicolumn{1}{c|}{\cellcolor[HTML]{75C47F}\textbf{{0.739}}} &
  \multicolumn{1}{c|}{\cellcolor[HTML]{E7E6E6}} \\ \cline{1-9}
\multicolumn{1}{|l|}{{LLaMa 2 7B}   } &
  \multicolumn{1}{c|}{\cellcolor[HTML]{8ACB84}0.627} &
  \multicolumn{1}{c|}{\cellcolor[HTML]{91CD85}\textbf{{0.587}}} &
  \multicolumn{1}{c|}{\cellcolor[HTML]{8CCB84}0.615} &
  \multicolumn{1}{c|}{\cellcolor[HTML]{90CD85}0.594} &
  \multicolumn{1}{c|}{\cellcolor[HTML]{8BCB84}0.621} &
  \multicolumn{1}{c|}{\cellcolor[HTML]{93CD85}0.580} &
  \multicolumn{1}{c|}{\cellcolor[HTML]{95CE86}0.568} &
  \multicolumn{1}{c|}{\cellcolor[HTML]{E0E696}0.169} &
  \multicolumn{1}{c|}{\multirow{-4}{*}{\cellcolor[HTML]{E7E6E6}\emph{\begin{tabular}[c]{@{}c@{}}\end{tabular}}}} \\ \hline
\multicolumn{1}{|l|}{\textbf{FT of MLMs}} &
  \multicolumn{9}{c|}{\cellcolor[HTML]{E7E6E6}Fine-Tune of \textbf{Encoder-Only Models} (MLMs)} \\ \hline
\multicolumn{1}{|l|}{XLM-T} &
  \multicolumn{1}{c|}{\cellcolor[HTML]{E7E6E6}} &
  \cellcolor[HTML]{F3EC9A}0.067 &
  \cellcolor[HTML]{EAE998}0.113 &
  \cellcolor[HTML]{FFEF9C}0.001 &
  \cellcolor[HTML]{FAEE9B}0.029 &
  \cellcolor[HTML]{FFEF9C}0.000 &
  \cellcolor[HTML]{F3EC9A}0.067 &
  \cellcolor[HTML]{F5EC9A}0.054 &
  \cellcolor[HTML]{69C07D}\textbf{0.802} \\ \cline{1-1}
\multicolumn{1}{|l|}{XLM-R} &
  \multicolumn{1}{c|}{\cellcolor[HTML]{E7E6E6}} &
  \cellcolor[HTML]{EBE998}0.107 &
  \cellcolor[HTML]{F3EC9A}0.067 &
  \cellcolor[HTML]{E7E897}0.130 &
  \cellcolor[HTML]{F4EC9A}0.062 &
  \cellcolor[HTML]{C2DC8F}0.328 &
  \cellcolor[HTML]{E7E897}0.133 &
  \cellcolor[HTML]{FFEF9C}0.001 &
  \cellcolor[HTML]{6BC17D}\textbf{{0.791}} \\ \cline{1-1}
\multicolumn{1}{|l|}{mBERT} &
  \multicolumn{1}{c|}{\multirow{-3}{*}{\cellcolor[HTML]{E7E6E6}}} &
  \cellcolor[HTML]{EAE998}0.115 &
  \cellcolor[HTML]{EBE998}0.108 &
  \cellcolor[HTML]{F0EB99}0.083 &
  \cellcolor[HTML]{FEEF9C}0.007 &
  \cellcolor[HTML]{FFEF9C}0.000 &
  \cellcolor[HTML]{FFEF9C}0.000 &
  \cellcolor[HTML]{FFEF9C}0.000 &
  \cellcolor[HTML]{66BF7C}\textbf{{0.818}} \\ \hline
\multicolumn{1}{|l|}{\textbf{{FT of LLMs}}} &
  \multicolumn{9}{c|}{\cellcolor[HTML]{E7E6E6}{Fine-Tune of \textbf{Encoder-Decoder and Decoder-Only Models} (LLMs)}} \\ \hline
\multicolumn{1}{|l|}{{LLaMa 2 7B}   } &
  \multicolumn{1}{c|}{\cellcolor[HTML]{E7E6E6}} &
  \multicolumn{1}{c|}{\cellcolor[HTML]{FFEF9C}0.000} &
  \multicolumn{1}{c|}{\cellcolor[HTML]{FFEF9C}0.000} &
  \multicolumn{1}{c|}{\cellcolor[HTML]{FFEF9C}0.000} &
  \multicolumn{1}{c|}{\cellcolor[HTML]{FFEF9C}0.000} &
  \multicolumn{1}{c|}{\cellcolor[HTML]{FFEF9C}0.000} &
  \multicolumn{1}{c|}{\cellcolor[HTML]{FFEF9C}0.000} &
  \multicolumn{1}{c|}{\cellcolor[HTML]{D5E293}0.228} &
  \multicolumn{1}{c|}{\cellcolor[HTML]{7CC681}\textbf{{0.701}}} \\ \cline{1-1} \cline{3-10} 
\multicolumn{1}{|l|}{{FlanT5} } &
  \multicolumn{1}{c|}{\multirow{-2}{*}{\cellcolor[HTML]{E7E6E6}}} &
  \multicolumn{1}{c|}{\cellcolor[HTML]{FFEF9C}0.000} &
  \multicolumn{1}{c|}{\cellcolor[HTML]{FFEF9C}0.000} &
  \multicolumn{1}{c|}{\cellcolor[HTML]{FFEF9C}0.000} &
  \multicolumn{1}{c|}{\cellcolor[HTML]{FFEF9C}0.000} &
  \multicolumn{1}{c|}{\cellcolor[HTML]{FFEF9C}0.000} &
  \multicolumn{1}{c|}{\cellcolor[HTML]{FFEF9C}0.000} &
  \multicolumn{1}{c|}{\cellcolor[HTML]{FFEF9C}0.000} &
  \multicolumn{1}{c|}{\cellcolor[HTML]{68C07D}\textbf{{0.806}}} \\ \hline
\multicolumn{1}{|l|}{\textbf{EntLM}} &
  \multicolumn{9}{c|}{\cellcolor[HTML]{E7E6E6}{\textbf{Template-Free Few-Shot} in \textbf{Sequence Labeling} Tasks for MLMs}} \\ \hline
\multicolumn{1}{|l|}{mBERT} &
  \multicolumn{1}{c|}{\cellcolor[HTML]{E7E6E6}} &
  \multicolumn{1}{c|}{\cellcolor[HTML]{C4DD90}0.317} &
  \multicolumn{1}{c|}{\cellcolor[HTML]{B7D98D}0.385} &
  \multicolumn{1}{c|}{\cellcolor[HTML]{AED68B}0.437} &
  \multicolumn{1}{c|}{\cellcolor[HTML]{9CD088}0.529} &
  \multicolumn{1}{c|}{\cellcolor[HTML]{96CE86}0.562} &
  \multicolumn{1}{c|}{\cellcolor[HTML]{91CD85}0.591} &
  \multicolumn{1}{c|}{\cellcolor[HTML]{92CD85}\textbf{{0.584}}} &
  \multicolumn{1}{c|}{\cellcolor[HTML]{6CC17D}0.788} \\ \hline
\multicolumn{1}{|l|}{\textbf{{GoLLIE}}} &
  \multicolumn{9}{c|}{\cellcolor[HTML]{E7E6E6}{{\textbf{Guideline following model} for Information Extraction}}} \\ \hline
\multicolumn{1}{|l|}{{GoLLIE 7B}} &
  \multicolumn{1}{c|}{\cellcolor[HTML]{82C882}0.670} &
  \multicolumn{1}{c|}{\cellcolor[HTML]{8BCB84}\textbf{{0.622}}} &
  \multicolumn{1}{c|}{\cellcolor[HTML]{89CA83}0.632} &
  \multicolumn{1}{c|}{\cellcolor[HTML]{83C982}0.662} &
  \multicolumn{1}{c|}{\cellcolor[HTML]{84C982}0.661} &
  \multicolumn{1}{c|}{\cellcolor[HTML]{7DC781}0.694} &
  \multicolumn{1}{c|}{\cellcolor[HTML]{7EC781}0.689} &
  \multicolumn{1}{c|}{\cellcolor[HTML]{76C47F}0.732} &
  \multicolumn{1}{c|}{\cellcolor[HTML]{63BE7B}\textbf{{0.832}}} \\ \hline
\end{tabular}}
\caption{{Named Entity Recognition (NER) for Locations with $k$-shot Sampling - Results on  Sequence Labeling Techniques (results in {\textbf{bold}} are referenced in the text)}}
\label{tab:resultNerTab}
\end{table}

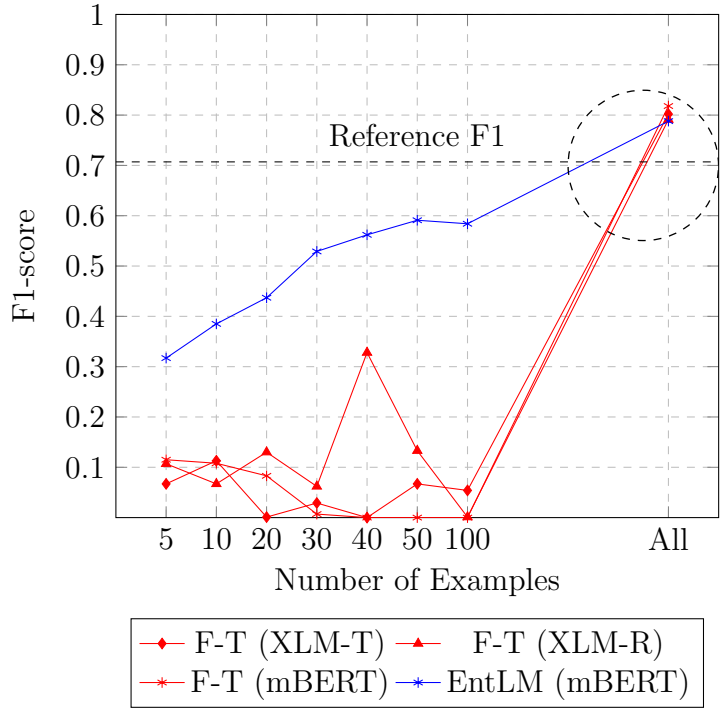
\begin{figure}[H]
\centering
\begin{tikzpicture}
\begin{axis}[
    xlabel={Number of Examples},
    ylabel={F1-score},
    xmin=0, xmax=100,
    ymin=0, ymax=0.9,
    legend cell align={left},
    ytick={0.1,0.2,0.3,0.4,0.5,0.6,0.7,0.8,0.9,1.0},
    xtick={10,20,30,40,50,60,70,100},
    xticklabels={5,10,20,30,40,50,100,All},       
    legend pos=south east,
    legend columns=1,
    ymajorgrids=true,
    xmajorgrids=true,
    grid style=dashed,
    width=1\columnwidth,
]
\addlegendimage{empty legend}

\addplot[
    color=ao,
    mark=diamond*,
]
    coordinates {
    (0,0.680)
    (10,0.704)
    (20,0.689)
    (30,0.730)
    (40,0.749)
    (50,0.741)
    (60,0.742)
    (70,0.739)

};

\addplot[
    color=blue,
    mark=diamond*,
]
    coordinates {
    (10,0.317)
    (20,0.385)
    (30,0.437)
    (40,0.529)
    (50,0.562)
    (60,0.591)
    (70,0.584)
    (100,0.788)

};

\addplot[
    color=red,
    mark=diamond*,
]
    coordinates {
    (10,0.115)
    (20,0.108)
    (30,0.083)
    (40,0.007)
    (50,0.000)
    (60,0.000)
    (70,0.000)
    (100,0.818)
};

\addplot[
    color=orange,
    mark=diamond*,
]
    coordinates {
    (10,0.000)
    (20,0.000)
    (30,0.000)
    (40,0.000)
    (50,0.000)
    (60,0.000)
    (70,0.000)
    (100,0.808)

};

\addplot[
    color=brown,
    mark=diamond*,
]
    coordinates {
    (0,0.670)
    (10,0.622)
    (20,0.632)
    (30,0.662)
    (40,0.661)
    (50,0.694)
    (60,0.689)
    (70,0.732)
    (100,0.832)

};

\draw [black, dashed, line width=0.5pt] (105,70) circle [radius=1cm];
\draw [black, dashed] (axis cs:0,0.707) -- (axis cs:120,0.707) node [pos=0.5, above, yshift=12pt] {Reference F1-score};

\addlegendentry{\hspace{0.8cm}\textbf{Techniques}}
\addlegendentry{{FS (Mistral 7B)}}
\addlegendentry{EntLM (mBERT)}
\addlegendentry{FT of MLMs (mBERT)}
\addlegendentry{{FT of LLMs (FlanT5)}}
\addlegendentry{{GoLLIE}}


\end{axis}
\end{tikzpicture}
\caption{{Named Entity Recognition (NER) for Locations -- $k$-shot Sampling}}
\label{fig:nerExample}
\end{figure}

{Figure~{\ref{fig:nerExample}} provides a chart view of the $k$-shot sampling results. For each technique in Table~{\ref{tab:resultNerTab}}, we report in the chart the results obtained with the most efficient open-source language model. Summarizing, several key takeaways can be drawn from Figure~{\ref{fig:nerExample}}:}

\begin{enumerate}
\item {With few examples, few-shot learning with LLMs, such as GPT 3.5, produces the by far the best results. It matches the \emph{word-matching} approach ({reference F1-score}) in zero-shot settings and surpasses it with as few as 5 shots. Therefore, it should be prioritized when working with limited examples in the tourism domain. LLMs, out of the box, possess substantial knowledge about locations, likely due to their recurrent exposure to similar tasks during training, which they can easily adapt. A few examples are sufficient to instruct these models on how to adapt this generic concept to domain-specific datasets. Alternatively, GoLLIE also achieves commendable results but only begins to surpass the rule-based approaches when provided with more than 50 examples per class. When using the full dataset, GoLLIE emerges as the optimal technique, achieving the highest performance and thus should be favored with a larger volume of examples.}
\item {Both fine-tuning techniques perform poorly in contexts with a low number of examples, as indicated by their low F1-scores. Traditionally, fine-tuning with a large dataset has been the preferred technique for NER for Locations, but the advent of generative language-based few-shot learning is beginning to shift this paradigm.
}
\item {EntLM with MLMs yields more modest results but remains viable in low-shot settings, unlike fine-tuning techniques.}
\end{enumerate}

\begin{figure}[H]
\centering
\begin{tikzpicture}
\begin{axis}[
    xlabel={\%  of Tweets},
    ylabel={F1-score},
    xmin=0, xmax=100,
    legend cell align={left},
    ymin=0, ymax=.9,
    ytick={0.1,0.2,0.3,0.4,0.5,0.6,0.7,0.8,0.9,1.0},
    xtick={5,10,20,30,40,50,60,70,80,90,100},
    xticklabels={5,10,20,30,40,50,60,70,80,90,100},       
    ymajorgrids=true,
    xmajorgrids=true,
    grid style=dashed,
    width=1\columnwidth,
    legend pos=south east,
    legend columns=1,
]
\addlegendimage{empty legend}
\addplot[
    color=blue,
    mark=asterisk,
]
    coordinates {
    (5,0.671)
    (10,0.686)
    (20,0.734)
    (30,0.752)
    (40,0.768)
    (50,0.771)
    (60,0.783)
    (70,0.779)
    (80,0.778)
    (90,0.791)
    (100,0.788)
};

\addplot[
    color=red,
    mark=diamond*,
]
    coordinates {
    (5,0.0)
    (10,0.0)
    (20,0.643)
    (30,0.705)
    (40,0.719)
    (50,0.748)
    (60,0.767)
    (70,0.788)
    (80,0.798)
    (90,0.790)
    (100,0.808)
};

\addplot[
    color=red,
    mark=triangle*,
]
    coordinates {
    (5,0.0)
    (10,0.424)
    (20,0.680)
    (30,0.744)
    (40,0.757)
    (50,0.781)
    (60,0.752)
    (70,0.791)
    (80,0.799)
    (90,0.805)
    (100,0.821)
};

\addplot[
    color=red,
    mark=asterisk,
]
    coordinates {
    (5,0.631)
    (10,0.692)
    (20,0.708)
    (30,0.765)
    (40,0.792)
    (50,0.815)
    (60,0.816)
    (70,0.833)
    (80,0.836)
    (90,0.831)
    (100,0.848)
};

\addplot[
    color=orange,
    mark=diamond*,
]
    coordinates {
    (5,0.5345)
    (10,0.5968)
    (20,0.6713)
    (30,0.6992)
    (40,0.6695)
    (50,0.6706)
    (60,0.6794)
    (70,0.7396)
    (80,0.6624)
    (90,0.7038)
    (100,0.701)
};

\addplot[
    color=orange,
    mark=triangle*,
]
    coordinates {
    (5,0.000)
    (10,0.3381)
    (20,0.5623)
    (30,0.6515)
    (40,0.6976)
    (50,0.7358)
    (60,0.7798)
    (70,0.7917)
    (80,0.7964)
    (90,0.8001)
    (100,0.806)
};

\draw [black,dashed] (axis cs:0,0.707) -- (axis cs:120,0.707) node [pos=0.5, below, yshift=0pt] {Reference F1-score};

\addlegendentry{\hspace{0.8cm}\textbf{Techniques}}
\addlegendentry{EntLM (mBERT)}
\addlegendentry{FT of MLMs (XLM-T)}
\addlegendentry{FT of MLMs (XLM-R)}
\addlegendentry{FT of MLMs (mBERT)}
\addlegendentry{{FT of LLMs (LLaMa 2 7B)}}
\addlegendentry{{FT of LLMs (FlanT5)}}

\label{fig:nerExamplePercentage}
\end{axis}

\end{tikzpicture}
\caption{{Named Entity Recognition (NER) for Locations -- Percentage Sampling}}
\label{fig:nerPercent}
\end{figure}
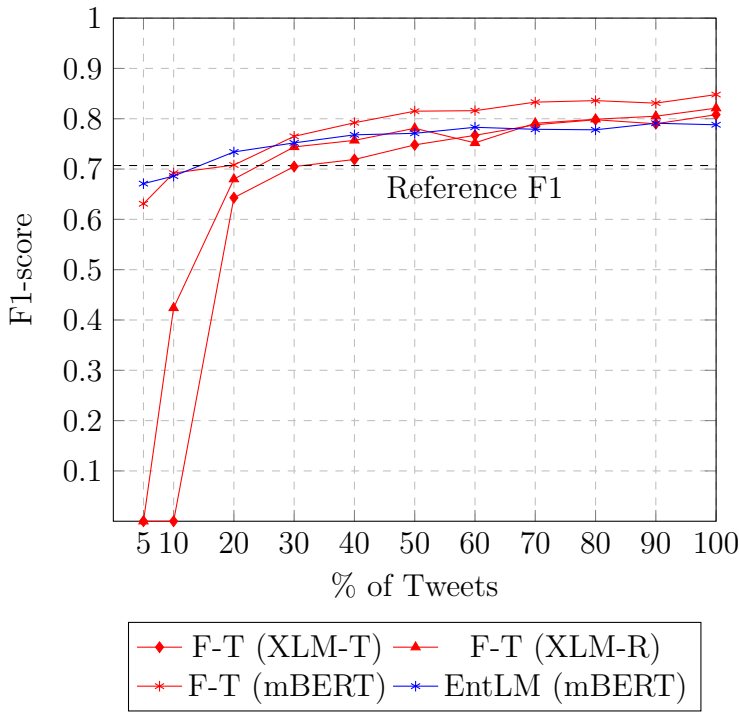

{Figure~\ref{fig:nerPercent} shows selected techniques used with percentage sampling (LLMs did not fit due to context constraints). In contrast to k-shot, it appears that percentage sampling is a better technique to set up EntLM and FT methods for sequence labeling. Thus, while EntLM is better with low amounts of data (5-10\%), the fine-tuned models exhibit a similar upward trend with as little as 10\% of the tweets, ending up outperforming EntLM as the number of data increases. We believe that the way this task is set up is perhaps not the best fit for EntLM, as the objective is to classify only one class, {location}, and EntLM's approach is based on generating {label words} which are associated with each entity class for better learning in few-shot settings. }

{However, this means that with only one class as target, all the label words are assigned to the same class, generating a noisy signal which, ultimately, as the number of words increases, hinders EntLM's performance. An interesting finding of our experiments (see Figure~{\ref{fig:nerPercent}}) is that both fine-tuned mBERT and EntLM are superior to the \emph{word-matching} algorithm when using only approximately 1000 tweets for training.}

{As depicted in Figure~{\ref{fig:nerPercent}}, approximately 20\% of the dataset (equivalent to about 330 tweets in our study) is necessary to achieve competitive results using this method. Marginally inferior outcomes are observed with Fine-tuning of LLMs, where LLaMa 2 7B slightly lags behind other models with an F1-score of {{0.701}}, while FlanT5 aligns with the MLM fine-tuning results, achieving an F1-score of {{0.806}}.}

{The EntLM technique shows that reliable results can be attained with less labeled data. Specifically, the fine-tuned mBERT does not surpass the performance of EntLM until more than 30\% of the training data is used (as shown in Figure~{\ref{fig:nerPercent}}, 30\%). Generally, fine-tuning becomes competitive only with percentage sampling, possibly due to the reduced frequency of O tokens in $k$-shot sampling. Conversely, EntLM demonstrates better performance with the same limited number of examples (as shown in Table~{\ref{tab:resultNerTab}}, {{0.584}} with 100 examples).} 

{Nevertheless, in scenarios with limited training examples, the performance of EntLM is not optimal. In such cases, our results indicate that prompt-based few-shot learning techniques with LLMs excel (refer to Table~{\ref{tab:resultNerTab}}, \emph{Prompt-based Few-Shot}). Particularly in zero-shot settings, the best results are achieved with GPT 4, which produces a remarkable F1-score of 0.829, equaling or surpassing all other strategies, even those involving extensive annotated datasets. Additionally, open-source (e.g., Mistral 7B, LLaMa 2 7B) or more cost-effective alternatives (e.g., GPT 3.5) also deliver satisfactory results in zero or few-shot settings (e.g., {{0.694}} for GPT 3.5, {{0.680}} for Mistral 7B in zero-shot settings), although they do not outperform fine-tuning techniques using the complete dataset.}

{Summarizing, optimal results in this task are achieved using Mistral with only 30 to 50 examples or, if computing requirements are too costly, with EntLM trained on 20\% of the data, which amounts to 300 tweets.}


{Having presented these findings, we now turn to the task of Fine-grained Thematic Concept Extraction.}

\subsection{Fine-grained Thematic Concept Extraction}

\begin{figure}[H]
\centering
\begin{tikzpicture}
\begin{axis}[
    xlabel={Number of Examples},
    ylabel={F1-score},
    xmin=0, xmax=120,
    ymin=0, ymax=.9,
    legend pos=south west,
    legend columns=1,
    legend cell align={left},
    ytick={0.1,0.2,0.3,0.4,0.5,0.6,0.7,0.8,0.9,1.0},
    xtick={10,20,30,40,50,60,70,110},
    xticklabels={5,10,20,30,40,50,100,All},       
    ymajorgrids=true,
    xmajorgrids=true,
    grid style=dashed,
    width=1\columnwidth,
]
\addlegendimage{empty legend}

\addplot[
    color=red,
    mark=diamond*,
]
    coordinates {
    (10,0.0)
    (20,0.0)
    (30,0.0)
    (40,0.0)
    (50,0.0)
    (60,0.0)
    (70,0.192)
    (110,0.309)
};

\addplot[
    color=red,
    mark=triangle*,
]
    coordinates {
    (10,0.0)
    (20,0.0)
    (30,0.0)
    (40,0.0)
    (50,0.0)
    (60,0.0)
    (70,0.024)
    (110,0.249)
};

\addplot[
    color=red,
    mark=asterisk,
]
    coordinates {
    (10,0.0)
    (20,0.0)
    (30,0.0)
    (40,0.0)
    (50,0.0)
    (60,0.0)
    (70,0.157)
    (110,0.201)
};

\addplot[
    color=blue,
    mark=asterisk,
]
    coordinates {
    (10,0.760)
    (20,0.790)
    (30,0.805)
    (40,0.814)
    (50,0.821)
    (60,0.824)
    (70,0.833)
    (110,0.838)

};
\draw [black,dashed] (axis cs:0,0.836) -- (axis cs:120,0.836) node [pos=0.5, above, yshift=2pt] {Reference F1-score};


\addlegendentry{\hspace{0.8cm}\textbf{Techniques}}
\addlegendentry{FT of MLMs (XLM-T)}
\addlegendentry{FT of MLMs (XLM-R)}
\addlegendentry{FT of MLMs (mBERT)}
\addlegendentry{EntLM (mBERT)}
\end{axis}
\end{tikzpicture}

\caption{Fine-grained Thematic Concept Extraction -- $k$-shot Sampling}
\label{fig:conceptExample}
\end{figure}
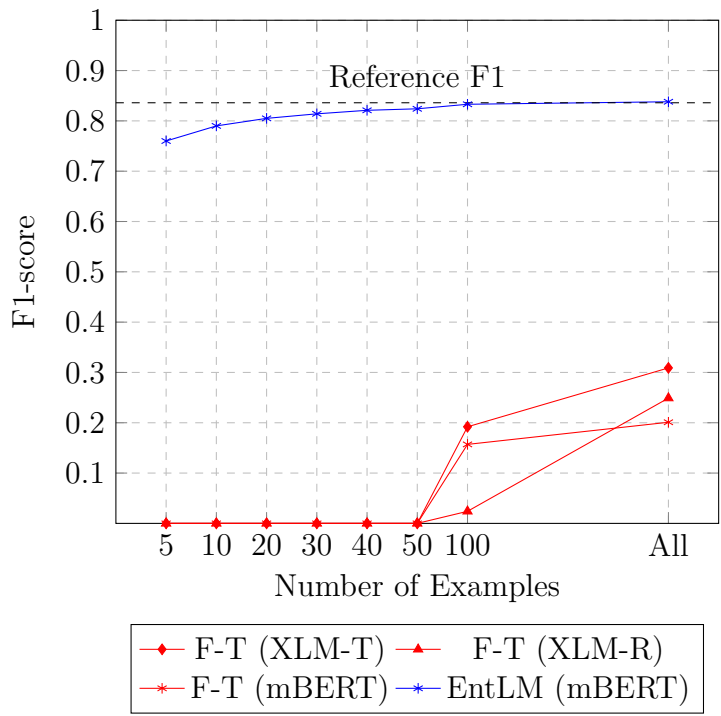

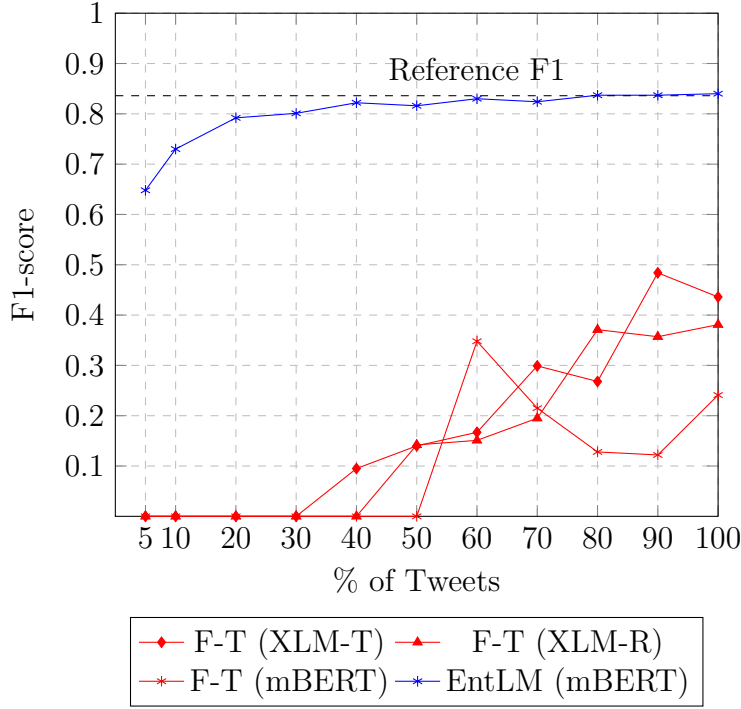
\begin{figure}[H]
\centering
\begin{tikzpicture}
\begin{axis}[
    xlabel={\%  of Tweets},
    ylabel={F1-score},
    xmin=0, xmax=100,
    ymin=0, ymax=.9,
    ytick={0.1,0.2,0.3,0.4,0.5,0.6,0.7,0.8,0.9,1.0},
    xtick={5,10,20,30,40,50,60,70,80,90,100},
    xticklabels={5,10,20,30,40,50,60,70,80,90,100},       
    legend style={at={(0.27,0.5)},
    anchor=north,legend columns=1},
    legend cell align={left},
    legend columns=1,
    ymajorgrids=true,
    xmajorgrids=true,
    grid style=dashed,
    width=1\columnwidth,
]
\addlegendimage{empty legend}

\addplot[
    color=red,
    mark=diamond*,
]
    coordinates {
    (5,0.0)
    (10,0.0)
    (20,0.0)
    (30,0.0)
    (40,0.095)
    (50,0.140)
    (60,0.167)
    (70,0.299)
    (80,0.268)
    (90,0.484)
    (100,0.436)
};

\addplot[
    color=red,
    mark=triangle*,
]
    coordinates {
    (5,0.0)
    (10,0.0)
    (20,0.0)
    (30,0.0)
    (40,0.0)
    (50,0.142)
    (60,0.151)
    (70,0.195)
    (80,0.371)
    (90,0.357)
    (100,0.381)
};

\addplot[
    color=red,
    mark=asterisk,
]
    coordinates {
    (5,0.0)
    (10,0.0)
    (20,0.0)
    (30,0.0)
    (40,0.0)
    (50,0.0)
    (60,0.348)
    (70,0.215)
    (80,0.128)
    (90,0.122)
    (100,0.241)
};

\addplot[
    color=blue,
    mark=asterisk,
]
    coordinates {
    (5,0.648)
    (10,0.730)
    (20,0.792)
    (30,0.801)
    (40,0.822)
    (50,0.816)
    (60,0.830)
    (70,0.824)
    (80,0.837)
    (90,0.837)
    (100,0.840)
};
\draw [black,dashed] (axis cs:0,0.836) -- (axis cs:120,0.836) node [pos=0.5, above, yshift=2pt] {Reference F1-score};
\addlegendentry{\hspace{0.8cm}\textbf{Techniques}}
\addlegendentry{FT of MLMs (XLM-T)}
\addlegendentry{FT of MLMs (XLM-R)}
\addlegendentry{FT of MLMs (mBERT)}
\addlegendentry{EntLM (mBERT)}

\end{axis}
\end{tikzpicture}
\caption{Fine-grained Thematic Concept Extraction -- Percentage Sampling}
\label{fig:conceptPercent}
\end{figure}

Perhaps it is in the evaluation of {Fine-grained Thematic Concept Extraction}, shown in Figures~\ref{fig:conceptExample} and \ref{fig:conceptPercent}, where few-shot learning with MLMs for sequence labeling clearly makes its mark. For a task that involves detecting and classifying sequences into a predetermined inventory of 315 classes, EntLM paired with mBERT performs very competitively. Thus, with just five examples per class (5-shot setting), it obtains a 0.760 F1-score, almost equaling the \emph{word-matching} algorithm's results with just a 50-shot training. These scores indicate a strong ability to accurately identify touristic concepts, as reflected by the high precision values spanning from 0.80 to 0.91.

Although overall results with \emph{word-matching} were similar, EntLM was slightly superior in terms of recall while being slightly worse in precision. Still, EntLM's performance shows great promise to avoid costly manual annotation effort or complex development of rule-based algorithms for domain-specific fine-grained sequence labeling tasks. EntLM's results are perhaps magnified by the very poor results obtained by the fine-tuning techniques in both data sampling scenarios, which indicates the difficulty of learning good sequence taggers for fine-grained tasks.

Let's now move on to case studies illustrating the difficulty of the tasks.

\subsection{{Case Study: Visualization of the Results for the Tourism Domain}}

\begin{figure}[H]
    \centering
    \includegraphics[width=1\linewidth]{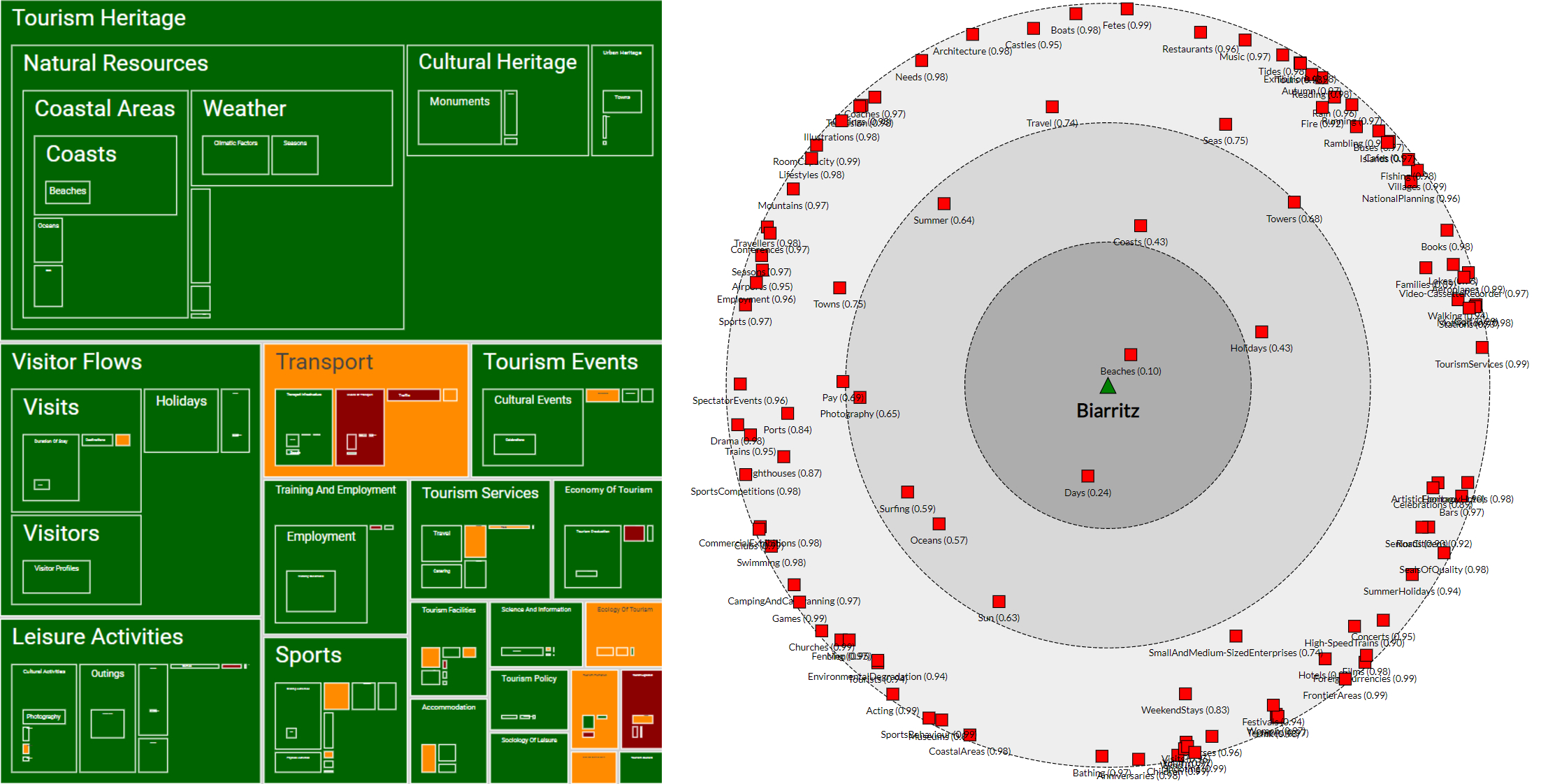}
    \caption{{Example of NLP Results Usage -- Tourism Thematic Map with Sentiment Overlay ({left}) and Proxemic View ({right})}}
    \label{fig:visu}
\end{figure}

{Figure{~\ref{fig:visu}} shows a visualization of the results generated using a dashboard that was created as part of our project to visualize NLP annotations, see {\cite{masson2024textbi}} for more details. Here, we have loaded the annotations generated by the best techniques for each task.}

{Firstly, on the left, we have a thematic map (represented as a multilevel treemap). Here, each square represents a thematic concept, and the size of each square represents the proportion of the concept in the dataset. As we can see, \emph{TourismHeritage}-related concepts are in the majority (e.g., \emph{NaturalResources}, \emph{CulturalHeritage}). Additionally, we have superimposed a sentiment overlay on the thematic map to visualize the aggregated sentiment for each thematic concept. Here, green signifies that the sentiment tends to be positive, orange mixed, and red negative. We can observe that at the scale of our dataset, most touristic concepts tend to be associated with a positive sentiment, except some branches like \emph{Transportation}, which are more mixed and even plainly negative for some concepts.}

{On the right is a proxemic crosshair {\cite{hall1966hidden}}. Here, a location named entity (the city of \emph{Biarritz}) is represented at the center, and the most co-occurring thematic concepts are scattered around it. This allows us to visualize for a given location, which touristic concepts it is mostly associated with. In this case, \emph{Biarritz} is mostly associated with the concepts of \emph{Beaches}, \emph{Holidays}, \emph{Coasts}, \emph{Days}, and \emph{Ocean}.}

{This visualization illustrates the potential of this kind of NLP annotations for stakeholders in the tourism domain (e.g., tourism offices, local municipalities, etc.). Having conducted these experiments, we will now delve into the insights gained and discuss the key findings that emerged. Additionally, we will address the potential limitations and biases that might have affected the results. }

\section{Discussion and {and Error Analysis}}\label{sec:discussion}

In our comparative analysis of various techniques for three knowledge extraction tasks in the tourism domain, namely Sentiment Analysis, NER for Locations, and Fine-grained Thematic Concept Extraction, we have
gained interesting insights into the data requirements and performance of these techniques.

\subsection{Sentiment Analysis}
For {Sentiment Analysis}, our findings suggest that a model previously fine-tuned on a large out-of-domain dataset for the same downstream text classification task can outperform prompt-based techniques even in few-shot settings. If such extra data is not available for our target task, {then addressing the task by means of few-shot learning techniques with LLMs (in particular the GPT series of models and Mistral 7B) has been demonstrated to be the best technique to avoid costly manual annotation work}.

{As for misclassification cases, they often arise from the model's reliance on surface-level lexical features without adequately accounting for context or tone. For example, the tweet ``\textit{Feu d’artifice du 14 juillet à Hendaye}''\footnote{English version: ``July 14th fireworks in Hendaye''.}, referring to a national celebration, is incorrectly labeled as negative by XLM-T Sentiment despite its neutral or festive nature. Similarly, ``\textit{Balade à Hendaye \#plage \#mer \#architecture}''\footnote{English version: ``Walk in Hendaye \#beach \#sea \#architecture''.} describes a leisure activity in a neutral setting, but is misclassified, likely due to the absence of explicit sentiment words and overemphasis on structural patterns.}

\subsection{Named Entity Recognition (NER) for Locations}

{Regarding {NER for Locations}, the optimal strategy would be to use Mistral in few-shot with only 30 examples. Failing that, EntLM performs well when trained on percentage sampling using 20\% of the training data.}

The case of EntLM is interesting because, in this task, there is only a single class but with many label words associated with it (995 different location names). We hypothesized that associating many label words with the same class does not benefit EntLM. 

{Interestingly, prompt-based few-shot learning with LLMs and GoLLIE generally performs exceptionally well in contexts with limited examples. This effectiveness is largely due to their design, which leverages extensive pre-training on diverse data, likely containing a lot of location entities, allowing them to generalize this task from minimal input. Both the GPT series of models and Mistral 7B, for instance, have demonstrated very good results in these scenarios. Their ability to adapt quickly with little to no additional training data makes them the best techniques for zero-shot or few-shot settings.}

{Regarding the misclassification cases in NER for Locations, they frequently result from ambiguity in sequence boundaries, improper capitalization, or domain-specific phrasing that deviates from standard named entity patterns. For instance, a phrase like ``congrès \#sniteat-unsa à Hendaye''\footnote{English version: ``conference \#sniteat-unsa in Hendaye''.} may lead the models to misidentify ``\#sniteat-unsa'' as a location rather than an organization (SNITEAT-UNSA is a French trade union) due to the hashtag and its syntactic position. Similarly, ``Plage d’Hendaye''\footnote{English version: ``Hendaye Beach''.} might be incorrectly labeled as a facility instead of a geographic location. These cases illustrate challenges in handling informal structures, social media conventions, and multilingual inputs.}

We also explored the idea of improving the fine-tuning process of our dataset by combining it with other existing corpora from other domains (see {Figure{~\ref{c3:fig:nerPercentAll}}}), such as AnCora \cite{ancora2008}
({Spanish}) and ESTER \cite{ester2006} ({French}), both
already annotated with location entities. However, this experiment did not lead to any significant improvements in the F1-score as shown in {Figure{~\ref{c3:fig:nerPercentAll}}}. Upon merging the three datasets, the F1-score saw only a minor increase, rising from 0.808 to 0.835 for XLM-T and from 0.821 to 0.830 for XLM-R. The limited
improvement could be attributed to the fact that these corpora are not specifically designed for social media. ESTER consists of radio broadcast transcripts, while AnCora consists of newspaper texts. Consequently, they lack the contextual information pertinent to the tourism domain.

\begin{figure}[H]
\centering
\begin{tikzpicture}
\begin{axis}[
    xlabel={\%  of Tweets},
    ylabel={F1-score},
    xmin=0, xmax=100,
    ymin=0, ymax=.9,
    legend columns=1,
    legend cell align={left},
    ytick={0.1,0.2,0.3,0.4,0.5,0.6,0.7,0.8,0.9,1.0},
    xtick={5,10,20,30,40,50,60,70,80,90,100},
    xticklabels={5,10,20,30,40,50,60,70,80,90,100},  
    legend style={at={(0.5,-0.2)},anchor=north},
    ymajorgrids=true,
    xmajorgrids=true,
    grid style=dashed,
    width=.9\textwidth,
]
\addlegendimage{empty legend}

\addplot[
    color=blue,
    mark=diamond*,
]
    coordinates {
    (5,0.0)
    (10,0.0)
    (20,0.643)
    (30,0.705)
    (40,0.719)
    (50,0.748)
    (60,0.767)
    (70,0.788)
    (80,0.798)
    (90,0.790)
    (100,0.808)
};

\addplot[
    color=blue,
    mark=triangle*,
]
    coordinates {
    (5,0.0)
    (10,0.424)
    (20,0.680)
    (30,0.744)
    (40,0.757)
    (50,0.781)
    (60,0.752)
    (70,0.791)
    (80,0.799)
    (90,0.805)
    (100,0.821)
};

\addplot[
    color=ao,
    mark=diamond*,
]
    coordinates {
    (5,0.566)
    (10,0.637)
    (20,0.689)
    (30,0.740)
    (40,0.768)
    (50,0.788)
    (60,0.807)
    (70,0.819)
    (80,0.822)
    (90,0.827)
    (100,0.835)
};

\addplot[
    color=ao,
    mark=triangle*,
]
    coordinates {
    (5,0.556)
    (10,0.620)
    (20,0.670)
    (30,0.728)
    (40,0.751)
    (50,0.779)
    (60,0.793)
    (70,0.810)
    (80,0.811)
    (90,0.815)
    (100,0.830)
};

\draw [black,dashed] (axis cs:0,0.707) -- (axis cs:120,0.707) node [pos=0.5, below, yshift=0pt] {Reference F1-score};


\addlegendentry{\hspace{2.6cm}\textbf{Techniques}}
\addlegendentry{{FT of MLMs (XLM-T, tourism dataset only)}}
\addlegendentry{{FT of MLMs (XLM-R, tourism dataset only)}}
\addlegendentry{{FT of MLMs (XLM-T, combined dataset)}}
\addlegendentry{{FT of MLMs (XLM-R, combined dataset)}}

\end{axis}

\end{tikzpicture}
\caption{{Comparison of Named Entity Recognition (NER) Model Performance: Our Dataset ({{{blue}}}) Compared to the Combined Dataset ({\textcolor{ao}{{green}}}) -- Fine-Tuning}}
\label{c3:fig:nerPercentAll}
\end{figure}

\subsection{Fine-grained Thematic Concept Extraction}
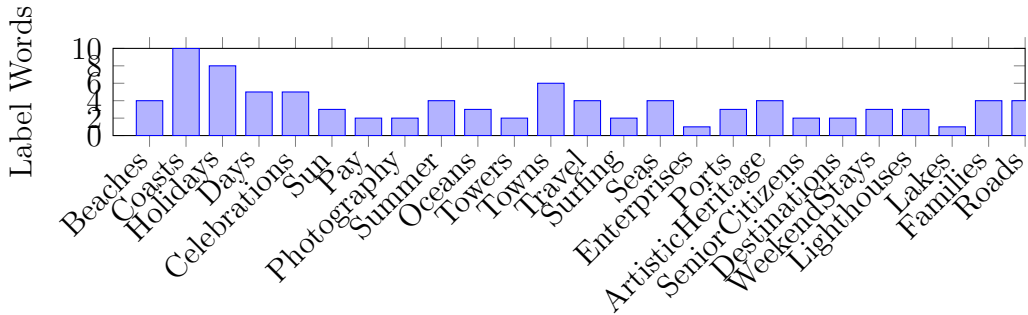
\begin{figure}[H]
\centering
\begin{tikzpicture}
    \begin{axis}[
        width=\textwidth,
        height=0.5\textwidth,
        ybar,
        ymin=0,
        ymax=10,
        xmin=0,
        xmax=25,
        xtick=data,
        xticklabels={Beaches, Coasts, Holidays, Days, Celebrations, Sun, Pay, Photography, Summer, Oceans, Towers, Towns, Travel, Surfing, Seas, Enterprises, Ports, ArtisticHeritage, SeniorCitizens, Destinations, WeekendStays, Lighthouses, Lakes, Families, Roads, Men, Rambling, NationalPlanning, Exhibitions, Concerts, Fire, High-SpeedTrains, Hotels, Trains, Bars, Drawing, Rain},
        xticklabel style={rotate=45, anchor=east},
        xlabel={},
        ylabel={Label Words}
    ]
    \addplot coordinates {
(1,4)
(2,10)
(3,8)
(4,5)
(5,5)
(6,3)
(7,2)
(8,2)
(9,4)
(10,3)
(11,2)
(12,6)
(13,4)
(14,2)
(15,4)
(16,1)
(17,3)
(18,4)
(19,2)
(20,2)
(21,3)
(22,3)
(23,1)
(24,4)
(25,4)
(26,4)
(27,3)
(28,2)
(29,3)
(30,5)
    };
    \end{axis}
\end{tikzpicture}
\caption{{Number of Label Words for the Most Frequent Thematic Concepts Found in the Tweets}}
\label{fig:labelWords}
\end{figure}

In the case of Fine-grained Thematic Concept Extraction, it is a different and more
complex sequence labeling task, which involves a large inventory of classes (315 concepts instantiated out of the
1,494 from the WTO tourism thesaurus), each having very few instance label words that are
highly representative of the classes to which they refer.
Figure~\ref{fig:labelWords} shows the low count of unique label words for the thematic concepts most often found in the tweets.
As we expected, fine-tuning did not yield any satisfactory results regardless of
the amount of data used or the sampling technique employed. We believe this is due
to the low number of examples per class. However, a major finding of this paper
is that the EntLM few-shot learning techniques demonstrated solid performance for
this task, even when trained with a limited number of examples on a very small
percentage of the available data. This robustness could be attributed to the
model's ability to generalize effectively from a smaller set of label words, which are associated with each of the classes, namely, the {thematic
concepts}. This result is particularly useful for practical applications where
manually annotated data is usually very scarce.

{We also performed a manual inspection of the misclassification cases in thematic concept detection, errors are largely due to the high number of distinct themes combined with sparse training instances per theme. This imbalance hinders the models' ability to generalize, leading to poor recognition of underrepresented concepts. Additionally, the models often fail to detect concepts when they are misspelled, abbreviated, or expressed in non-standard forms: e.g., ``adminstratif'' instead of ``administratif'' or hashtags like ``\#contentieuxadministratif''\footnote{English version: ``\#administrativelitigation''} being overlooked due to their concatenated structure. These factors reduce robustness in real-world, noisy text environments.}

Our experiments provide interesting insights but have some limitations. Firstly,
we intentionally focused on three common NLP tasks within the tourism domain,
ensuring an in-depth understanding of this area. Although our findings are specific
to these tasks, additional studies can broaden the applicability to other
domains. Second, we worked with a curated dataset of 2,961 tweets. This
limitation in dataset size was purposefully chosen to maintain focus, but
larger and more diverse datasets might offer further perspectives. Lastly,
while our dataset has a diverse language representation, it predominantly
features French tweets, mirroring the tourist demographics of the {French Basque
Coast} region.

\section{Concluding Remarks and Future Work}\label{sec:conclusion}

In this paper, we present a comparative study of several Natural Language Processing (NLP) strategies for Sentiment Analysis, NER for Locations, and {Fine-grained Thematic Concept Extraction} for social media data in the tourism domain. 

The main objective of this work is to establish the best approaches to obtain competitive results while keeping to a minimum any costly manual annotation or the development of complex and cumbersome rule-based approaches, especially for complex tasks such as {Fine-grained Thematic Concept Extraction}. In order to be able to do the experimentation, we provide a novel tourist-specific multilingual (French, English, and Spanish) dataset annotated for the tasks that are the subject of our study. This dataset, one of its kind, will be made publicly available in the future to facilitate further research on this particular topic and the reproducibility of the results.

{Results show that current few-shot learning techniques allow us to obtain competitive results for all three tasks with a very small amount of annotation: 5 tweets per label (15 in total) for Sentiment Analysis, 30 examples of Location for {NER}, and 1000 tweets annotated with {thematic concepts}. We believe that these findings are helpful not only for the development of our own application but also for other domain-specific applications, which may require NLP analysis as a model enrichment. However, further research on analogous NLP tasks in different domains would be needed to validate the generalizability of our results in various domains, other than tourism. We believe these findings offer value not only for NLP researchers focusing on tourism but also for applications needing domain-specific NLP analysis, especially when there is a lack of annotated data or when one wishes to exclude ad hoc rule-based approaches.}

Our project's next step involves presenting the NLP results to stakeholders in the tourism industry. To achieve this, we have initiated the development of multidimensional, dynamic dashboards based on the output generated by the NLP processing modules discussed in this paper. These dashboards are designed to highlight the frequency, relationships, and trajectories of extracted entities, categorized spatially (cities, POIs), temporally (date, time periods), and thematically (touristic concepts), within a specified social media corpus, all correlated with contextual data such as sentiment and engagement levels.

 \bibliographystyle{elsarticle-num} 
 \bibliography{cas-refs}





\end{document}